\documentclass[10pt,twocolumn,letterpaper]{article}

\usepackage{cvpr22/cvpr}              %
\usepackage[accsupp]{axessibility}  %

\usepackage{graphicx}
\usepackage{amsmath}
\usepackage{amssymb}
\usepackage[symbol]{footmisc}

\usepackage[pagebackref,breaklinks,colorlinks]{hyperref}
\usepackage[capitalize]{cleveref}
\crefname{section}{Sec.}{Secs.}
\Crefname{section}{Section}{Sections}
\Crefname{table}{Table}{Tables}
\crefname{table}{Tab.}{Tabs.}

\def\cvprPaperID{736} %
\def\confName{CVPR}
\def\confYear{2022}

\usepackage{overpic}
\usepackage{enumitem} %
\usepackage{overpic} %
\usepackage{color}
\usepackage{booktabs}

\usepackage{amsfonts}       %
\usepackage{nicefrac}       %
\usepackage{microtype}      %
\usepackage{comment}
\usepackage[dvipsnames]{xcolor}
\usepackage{mathalpha}
\usepackage{float}
\usepackage{cuted}
\usepackage{multirow}
\usepackage{multicol}
\usepackage{multibib}

\usepackage{tabularx}
\usepackage{mathtools}
\usepackage{booktabs}
\usepackage{nonfloat}
\usepackage{enumitem}%
\usepackage{verbatim}
\usepackage{subcaption}
\usepackage{amsfonts} %
\DeclareMathSymbol{\shortminus}{\mathbin}{AMSa}{"39}
\DeclareFontFamily{OT1}{pzc}{}
\DeclareFontShape{OT1}{pzc}{m}{it}{<-> s * [1.150] pzcmi8t}{}
\DeclareMathAlphabet{\mathpzc}{OT1}{pzc}{m}{it}
\usepackage[mathscr]{eucal}
\usepackage{comment}
\usepackage{soul}
\usepackage{xfakebold}
\hypersetup{colorlinks,linkcolor={blue},citecolor={blue},urlcolor={red}}  

\newcommand{\fbseries}{\unskip\setBold\aftergroup\unsetBold\aftergroup\ignorespaces}
\makeatletter
\newcommand{\setBoldness}[1]{\def\fake@bold{#1}}
\makeatother

\definecolor{cadmiumgreen}{rgb}{0.0, 0.42, 0.24}

\DeclareMathSymbol{\shortminus}{\mathbin}{AMSa}{"39}

\newcommand{\fig}[1]{Fig~\ref{fig:#1}}
\newcommand{\sect}[1]{Sect~\ref{sect:#1}}

\newcommand{\tab}[1]{Table~\ref{tab:#1}}

\newcommand{\eq}[1]{Eq. (\ref{eq:#1})}

\newcommand{\MethodName}{RigNeRF\xspace}

\newcommand{\xbf}{\mathbf{x}}
\newcommand{\xbfcan}{\mathbf{x}_{\text{can}}}

\newcommand{\gammabf}{\pmb{\gamma}}

\definecolor{turquoise}{cmyk}{0.65,0,0.1,0.3}
\definecolor{purple}{rgb}{0.65,0,0.65}
\definecolor{dark_green}{rgb}{0, 0.5, 0}
\definecolor{orange}{rgb}{0.8, 0.6, 0.2}
\definecolor{red}{rgb}{0.8, 0.2, 0.2}
\definecolor{darkred}{rgb}{0.6, 0.1, 0.05}
\definecolor{blueish}{rgb}{0.0, 0.3, .6}
\definecolor{light_gray}{rgb}{0.7, 0.7, .7}
\definecolor{pink}{rgb}{1, 0, 1}
\definecolor{greyblue}{rgb}{0.25, 0.25, 1}
\definecolor{drivevid}{RGB}{75,154,250}
\definecolor{meshrig}{RGB}{241,135,40}
\definecolor{nvs}{RGB}{236,71,250}

\definecolor{boldblue}{RGB}{70, 118, 172}
\definecolor{boldgreen}{RGB}{40, 116, 12}

\usepackage{blindtext}

\renewcommand{\paragraph}[1]{\vspace{1em}\noindent\textbf{#1}.}
\usepackage{environ}
\NewEnviron{smequation}{%
    \begin{equation}
    \scalebox{0.85}{$\BODY$}
    \end{equation}
}

\begin{document}
\title{RigNeRF: Fully Controllable Neural 3D Portraits}

\author{
 ShahRukh Athar* \\
  Stony Brook University \\
  \texttt{sathar@cs.stonybrook.edu}
  \and
  Zexiang Xu \\
  Adobe Research \\
  \texttt{zexu@adobe.com} \\
  \and
  Kalyan Sunkavalli \\
  Adobe Research \\
  \texttt{sunkaval@adobe.com} \\
  \and
  Eli Shechtman\\
  Adobe Research \\
  \texttt{elishe@adobe.com} \\
  \and
  Zhixin Shu \\
  Adobe Research \\
  \texttt{zshu@adobe.com} \\
}

\maketitle
\begin{abstract}
\vspace{-1em}
Volumetric neural rendering methods, such as neural radiance fields (NeRFs), have enabled photo-realistic novel view synthesis. However, in their standard form, NeRFs do not support the editing of objects, such as a human head, within a scene. In this work, we propose \MethodName, a system that goes beyond just novel view synthesis and enables full control of head pose and facial expressions learned from a single portrait video. We model changes in head pose and facial expressions using a deformation field that is guided by a 3D morphable face model (3DMM). The 3DMM effectively acts as a prior for RigNeRF that learns to predict only residuals to the 3DMM deformations and allows us to render novel (rigid) poses and (non-rigid) expressions that were not present in the input sequence. Using only a smartphone-captured short video of a subject for training, we demonstrate the effectiveness of our method on free view synthesis of a portrait scene with explicit head pose and expression controls. 
The project page can be found \href{http://shahrukhathar.github.io/2022/06/06/RigNeRF.html}{here}.

\end{abstract}
\footnotetext[1]{Work done while interning at Adobe Research.}

\section{Introduction}
\label{sec:intro}

Photo-realistic editing of human portraits is a long-standing topic in the computer graphics and computer vision community. It is desirable to be able to control certain attributes of a portrait, such as 3D viewpoint, lighting, head pose, and even facial expression, after capturing. It also has great potential in AR/VR applications where a 3D immersive experience is valuable. However, it is a challenging task:
modeling and rendering a realistic human portrait with complete control over 3D viewpoint, facial expressions, and head pose in natural scenes remains elusive, despite the longtime interest and recently increased research.

3D Morphable Face Models (3DMMs) \cite{blanz1999morphable} were among the earliest attempts towards a fully controllable 3D human head model. 3DMMs use a PCA-based linear subspace to control face shape, facial expressions, and appearance independently. A face model of desired properties can be rendered in any view using standard graphics-based rendering techniques such as rasterization or ray-tracing. However, directly rendering 3DMMs \cite{blanz1999morphable}, which only models face region, is not ideal for photo-realistic applications as its lacks essential elements of the human head such as hair, skin details, and accessories such as glasses. Therefore, it is better employed as an intermediate 3D representation \cite{dnr, Kim2018DeepVP, tewari2020stylerig} 
due to its natural disentanglement of face attributes such as shape, texture, and expression, which makes 3DMMs an appealing representation for gaining control of face synthesis.

On the other hand, recent advances on neural rendering and novel view synthesis~\cite{nerf,nerfies,NerFACE, SRF, NeuralSceneFlow,DNeRF,  kaizhang2020,  Gao-portraitnerf,Liu-2020-NSV, Zhang-2020-NAA,Bemana-2020-XIN,Martin-2020-NIT,xian2020space} 
have demonstrated impressive image-based rendering of complex scenes and objects. Despite that, existing works are unable to simultaneously generate high quality novel views of a given natural scene and control the objects within it, including that of the human face and its various attributes. In this work, we would like to introduce a system to model a fully controllable portrait scene: with camera view control, head pose control, as well as facial expression control.

Control of head-pose and facial expressions can be enabled in NeRFs via a deformation module as done in \cite{DNeRF, nerfies, park2021hypernerf}. However, since those deformations are learnt in a latent space, they cannot be explicitly controlled.  A natural way add control to head-pose and facial expressions, via deformations, is by parameterizing the deformation field using the 3DMM head-pose and facial expression space. 
However, as shown in \fig{Hypernerfpp_comp}, such naive implementation of a deformation field leads to artefacts during the reanimation due to the loss of rigidity and incorrect modelling of facial expressions. 
To address these issue, we introduce \MethodName, a method that leverages a 3DMM to generate a coarse deformation field which is then refined by corrective residual predicted by an MLP to account for the non-rigid dynamics, hair and accessories. Beyond giving us a controllable deformation field, the 3DMM acts as an inductive bias allowing our network to generalize to \emph{novel} head poses and expressions that were not observed in the input video.

Our model is designed to be trained on a short video captured using a mobile device. Once trained, \MethodName allows for explicit control of head pose, facial expression and camera viewpoint. Our results capture rich details of the scene along with details of the human head such as the hair, beard, teeth and accessories. Videos reanimated using our method maintain high fidelity to both the driving morphable model in terms of facial expression and head-pose and the original captured scene and human head. 

 In summary, our contributions in this paper are as follows: 1) We propose a neural radiance field capable of full control of the human head along with simultaneously modelling the full 3D scene it is in. 2) We experimentally demonstrate the loss of rigidity when dynamic neural radiance fields are reanimated. 3) We introduce a deformation prior that ensures rigidity of the human head during reanimation thus significantly improves its quality.

\section{Related works}
\MethodName is a method for full control of head pose, facial expressions, and novel view synthesis of 3D portrait scene. It is closely related to recent work on neural rendering, novel view synthesis, 3D face modeling, and controllable face generation.

\paragraph{Neural Scene Representations and Novel View Synthesis} \MethodName is related to recent work in neural rendering and novel view synthesis \cite{nerf,park2021hypernerf,nerfies,NerFACE, IDR, SRF, lassner2020pulsar, lombardi2021mixture, NeuralSceneFlow,DNeRF, SVS, kaizhang2020, unisurf, Gao-portraitnerf,sitzmann2019scene,Liu-2020-NSV, Zhang-2020-NAA,Bemana-2020-XIN,Martin-2020-NIT,xian2020space, Wizadwongsa2021NeX}. Neural Radiance Fields (NeRF) use a Multi-Layer Perceptron (MLP), \(F\), to learn a volumetric representation of a scene. For every 3D point and the direction from which the point is being viewed, \(F\) predicts its color and volume density. For any given camera pose, \(F\) is first evaluated densely enough throughout the scene using hierarchical volume sampling \cite{nerf}, then volume rendering is used to render the final image. \(F\) is trained by minimizing the error between the predicted color of a pixel and its ground truth value. While NeRFs are able to generate photo-realistic images for novel view synthesis, it is only designed for a static scene and is unable to represent scene dynamics. Specifically designed for dynamic portrait video synthesis, our approach not only models the dynamics of human faces, but also allows specific controls on the facial animation. 

\paragraph{Dynamic Neural Scene Representations} Although NeRF~\cite{nerf} is designed for a static scene, several works have attempted to extend it to model dynamic objects or scene. There is a line of work \cite{NeuralSceneFlow, li2021neural, DNeRF, xian2020space} that extend NeRF to dynamic scenes by providing as input a time component and along with it imposing temporal constraints either by using scene flow \cite{NeuralSceneFlow, xian2020space} or by using a canonical frame \cite{DNeRF}. Similarly, Nerfies \cite{nerfies} too work with dynamic scenes by mapping to a canonical frame, however it assumes that the movement is small. In \cite{park2021hypernerf}, authors build upon \cite{nerfies} and use an ambient dimension in order to model topological changes in the deformation field. The deformation fields in these approaches are conditioned on learnt latent codes without specific physical or semantic meaning, and therefore not controllable in an intuitive manner. \MethodName, similar to \cite{park2021hypernerf, nerfies}, models the  portrait video by mapping to a canonical frame but in addition also enables full parameterized control of head pose and facial expression.

\paragraph{Controllable Face Generation} 
Recent breakthroughs in Generative Adversarial Networks(GANs)\cite{goodfellow2014generative, pix2pix2016, CycleGAN2017, StyleGAN, Karras-2019-ASB, Karras-2020-AAI} have enabled high-quality image generation and manipulation. They also inspired a large collection of work \cite{shu2018deforming, NeuralFace2017, athar2020self, pumarola2020ganimation, StarGAN2018, starganv2, tewari2020stylerig, tewari2020pie, deng2020disentangled, kowalski2020config} focusing on face image manipulation and editing. However, majority of these work are intrinsically image-based and lack explicit 3D representation. Therefore it is challenging to enable high-quality view synthesis and 3D controls of the portraits such as large pose changes or extreme facial expressions. Another line of work \cite{Kim-2018-DVP, Doukas2021Head2HeadDF, head2head2020, facedet3d} made use of 3D Morphable Model as intermediate 3D face representation to reanimate face images/videos. While being able to model head poses with great detail, thanks to the disentangled representation in 3DMM, they often unable to perform novel view synthesis as they focus on face region but neglect the geometry or appearance of the scene. Similarly, NerFACE\cite{NerFACE} uses neural radiance fields to model a 4D face avatar and allows pose/expression control on the head. However, they assume a static background and fixed camera, thus cannot perform view synthesis of the person or the scene. In contrast, our method \MethodName provides full control over the head pose and facial expressions of the person captured in the portrait video while simultaneously being able to synthesize novel views of the 3D portrait scene.

\paragraph{Hybrid Representations} The photorealism of volumetric and implicit representations have encouraged works that combine them with classical representations in order improve reconstruction \cite{chatziagapi2021sider} or lend control over foreground \cite{gafni2020dynamic, liu2021neural, peng2021animatable}. In \cite{liu2021neural}, authors learn a deformation field along with a texture mapping to reanimate human bodies. Similarly, \cite{peng2021animatable} learns a 3D skinning field to accurately deform points according to the target pose. Neither \cite{liu2021neural} nor \cite{peng2021animatable} model the full 3D scene. In contrast, \MethodName models the whole 3D scene with full control of head-pose, facial expressions and viewing directions.
\section{\MethodName}
\begin{figure*}[h]
    \centering
    \includegraphics[width=0.95\linewidth]{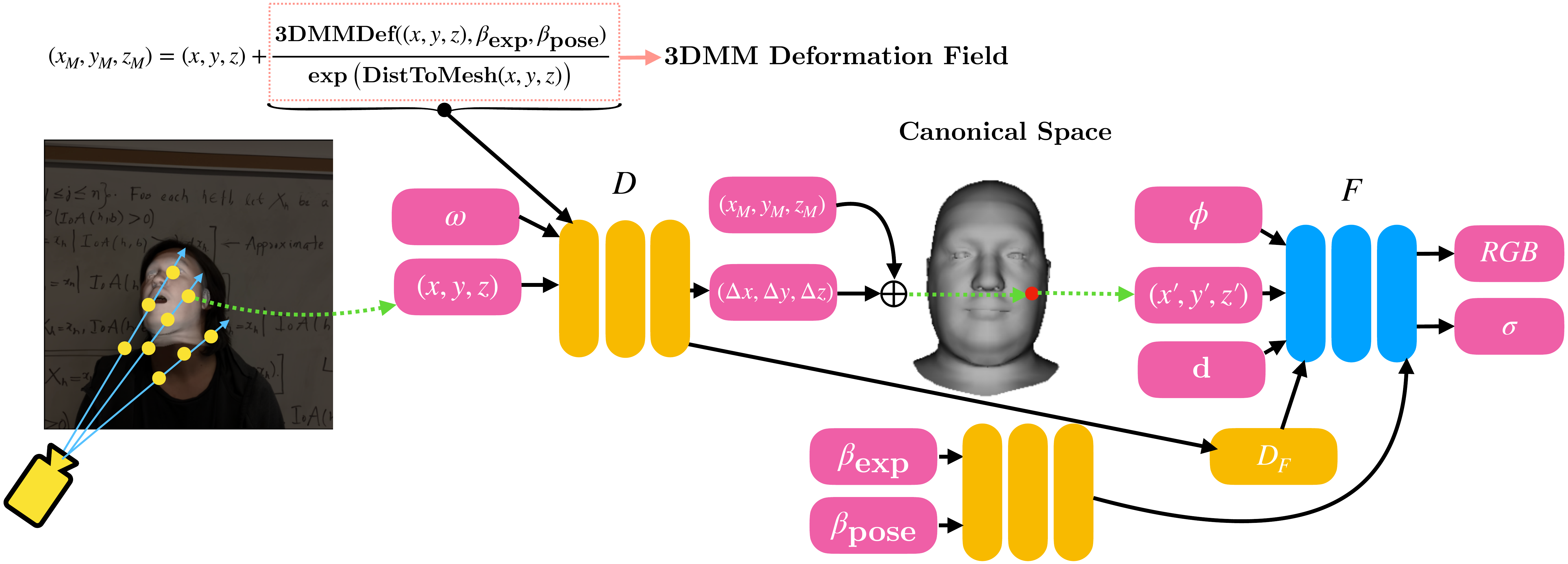}
    
    \caption{{\textbf{Overview of \MethodName.} \MethodName is a deformable NeRF architecture that consists of two learnable MLPs: a deformation MLP \(D\) and a color MLP \(F\). Given an image, we shoot rays through each of its pixels.
    For every ray, we deform each point on it according to a 3DMM-guided deformation field. This deformation field is the sum of the 3DMM deformation field  (see \sect{3dmmdef})  and the residual predicted by the deformation MLP, \(D\). Next, the deformed point is given as input to the color MLP, \(F\), which additionally takes as input the pose and expression parameters \(\{\beta_{\text{exp}}, \beta_{\text{pose}}\}\), the viewing direction \(\mathbf{d}\) and an appearance embedding \(\phi\) to predict the color and density. The final color of the pixel is calculated via volume rendering.
    }}
    \label{fig:method}
\end{figure*}

In this section, we describe our method, \MethodName, that enables novel view synthesis of 3D portrait scenes and arbitrary control of head pose and facial expressions. A Neural Radiance Field (NeRF) \cite{nerf} with a per-point deformation is used to control the head pose and facial expressions of the subject. The deformation field deforms the rays of each frame to a canonical space, defined by a the 3DMM in the frontal head-pose and neutral expression, where the colors are sampled. In order to model deformations due to both head-pose (a rigid deformation) and facial expressions (a non-rigid deformation) and to correctly deform facial details such as hair and glasses, the deformation field is defined as the sum of the 3DMM deformation field and a residual deformation predicted by a deformation MLP.

\subsection{Deformable Neural Radiance Fields}
A neural radiance field (NeRF) is defined as a continuous function \(F: \left(\gammabf_{m}(\mathbf{x}), \gammabf_{n}(\mathbf{d})\right) \rightarrow (\mathbf{c}(\xbf, \mathbf{d}), \sigma(\xbf))\), that, given the position of a point in the scene \(\mathbf{x}\) and the direction it is being viewed in, \(\mathbf{d}\), outputs the color \(\mathbf{c} = (r,g,b)\) and the density \(\sigma\). \(F\) is usually represented as a multi-layer perceptron (MLP) and \(\gammabf_{m}: \mathbb{R}^{3} \rightarrow \mathbb{R}^{3 + 6m}\) is the positional encoding \cite{nerf} defined as \(\gammabf_{m}(\xbf) = (\xbf,...,\text{sin}(2^{k}\xbf),\text{cos}(2^{k}\xbf),...)\) where \(m\) is the total number of frequency bands and \(k \in \{0,...,m-1\}\). The expected color of the pixel through which a camera ray 
passes is calculated via volume rendering. The parameters of \(F\) are trained to minimize the L2 distance between the expected color and the ground-truth.

NeRFs, as defined above, are designed for static scenes and offer no control over the objects within the scene. In order to model a dynamic scene, NeRFs are extended by additionally learning a deformation field to map each 3D point of the scene to a canonical space, where the volumetric rendering takes place \cite{nerfies, DNeRF, park2021hypernerf}. The deformation field is also represented by an MLP \(D_{i}: \xbf \rightarrow \xbfcan\) where  \(D_{i}\) is defined as \(D(\xbf, \omega_{i}) = \xbfcan\) and \(\omega_{i}\) is a per-frame latent deformation code. In addition to a deformation code, \(\omega_{i}\), a per-frame appearance code is also used \cite{nerfies, park2021hypernerf, DNeRF}, \(\phi_{i}\), thus the final radiance field for the \(i\)-th frame is as follows:
\begin{equation}
    (\mathbf{c}(\xbf, \mathbf{d}), \sigma(\xbf)) = F\left(\gammabf(D(\mathbf{x},\omega_{i})), \gammabf(\mathbf{d}), \phi_{i}\right)
    \label{eq:nerfie_def}
\end{equation}
In addition to the parameters of \(F\), each \(\omega_{i}\) and \(\phi_{i}\) are also optimized through stochastic gradient descent. While the aforementioned modifications are able generate novel views \cite{DNeRF, nerfies} of dynamic videos and handle small movement of objects in the scene \cite{nerfies}, the deformations they estimate are conditioned on learnt deformation codes that can be arbitrary. Instead, we seek intuitive deformation controls that explicitly disentangles and controls facial appearance based on camera viewpoint, head pose and expression.

\subsection{A 3DMM-guided deformation field}
\label{sect:3dmmdef}
\MethodName enables novel view synthesis of dynamic portrait scene and arbitrary control of head pose and facial expressions. For each frame \(i\), we first extract its head-pose and expression parameters \(\{\beta_{i,\text{exp}}, \beta_{i,\text{pose}}\}\) using DECA \cite{DECA} and landmark fitting \cite{3DDFA_V2}. Next, we shoot rays through each pixel, \(p\), of the frame and deform each point on the ray, \(\xbf\), to a position in the canonical space, \(\xbfcan = (x',y',z')\), where its color is computed.
A natural way to parameterize this canonical space, and any deviations from it, is using 3DMMs \cite{blanz1999morphable,FLAME:SiggraphAsia2017}. 
Thus, \MethodName's canonical space is defined as the one where the head has zero head-pose and a neutral facial expression. 

Unfortunately, a 3DMM is only defined accurately for a subset of points on the head---3DMM fitting is often not perfect and they do not model hair, glasses, etc.---and is undefined for point in the rest of 3D space. 
Hence, a deformation MLP \(D_{i}: \xbf \rightarrow \xbfcan\) is still necessary to perform the transformation to the canonical space. However, as detailed in \sect{hypernerfpp}, we find that directly predicting the deformation to the canonical space gives rise to artefacts during reanimation. The artefacts arise due to the inability of \(D\) to 1) maintain the rigidity of the head and 2) to model facial expressions correctly.

\begin{figure}[h]
   \hspace*{-0.6cm}
    \includegraphics[width=1.\linewidth]{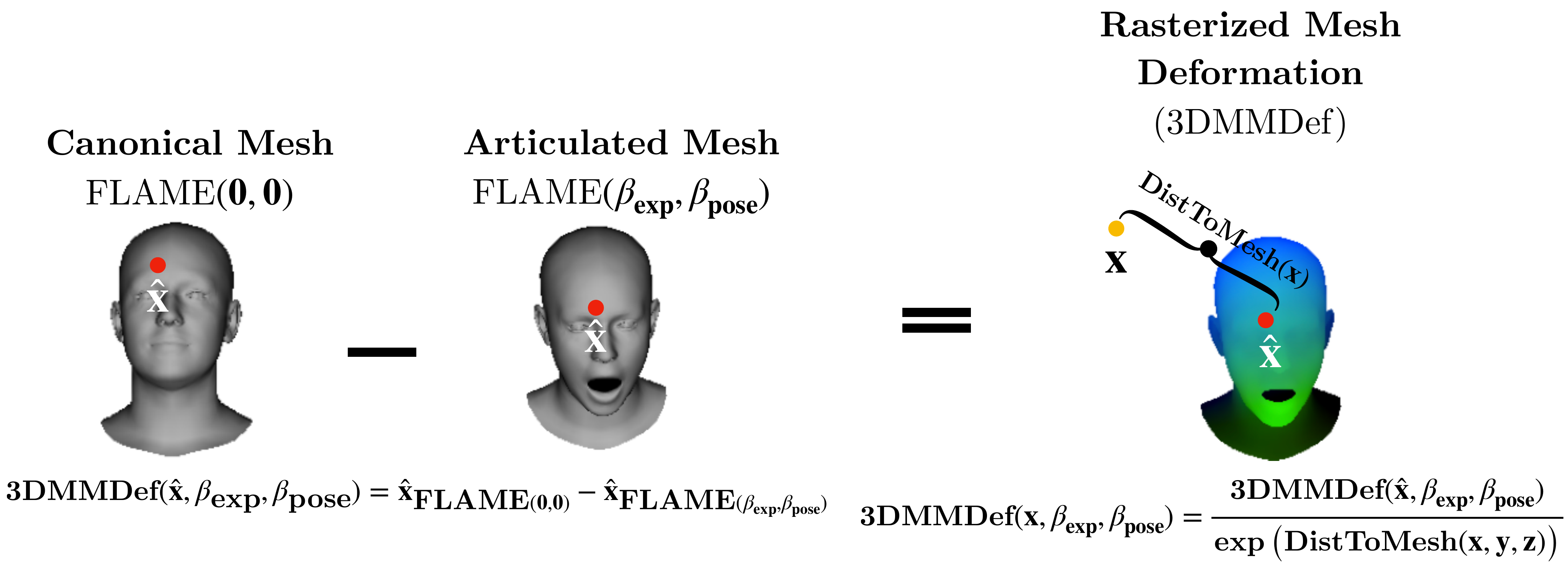}
    \caption{{\textbf{The 3DMM deformation field.} The 3DMM deformation field at any point in space, \(\xbf\), is equal to the deformation of its closest neighbor on the mesh, \(\hat{\xbf}\), weighted by the inverse of the exponential of the distance between \(\xbf\) and \(\hat{\xbf}\).
    }}
    \label{fig:meshdef}
\end{figure}

In order to fix this and ensure \MethodName is able to handle both rigid deformations due to head-pose changes and non-rigid deformations due to changes in facial expressions, we use a deformation field prior derived using the 3DMM. For expression and head-pose parameters, \(\{\beta_{\text{exp}}, \beta_{\text{pose}}\}\), the value of the 3DMM deformation field at any point \(\xbf = (x,y,z)\) is:

\begin{smequation}
    \text{3DMMDef}(\xbf, \beta_{\text{exp}}, \beta_{\text{pose}}) = \frac{\text{3DMMDef}(\hat{\xbf}, \beta_{\text{exp}}, \beta_{\text{pose}})}{\text{exp}(\text{DistToMesh}(\xbf))}
\end{smequation}
where, \(\text{3DMMDef}(\xbf) \) is the value of the 3DMM deformation field, \(\hat{\xbf} = (\hat{x},\hat{y},\hat{z})\) is the closest point to \((x,y,z)\) on the mesh and \(\text{DistToMesh} = ||\xbf - \hat{\xbf}||\) is the distance between \(\xbf\) and  \(\hat{\xbf}\). The 3DMM deformation of any point on the mesh, \(\hat{\xbf}\), is given by the difference between its position in the canonical space (i.e when the mesh had a zero head-pose and neutral facial expression) and its current articulation, as follows:
\begin{smequation}
    \text{3DMMDef}(\hat{\xbf},\beta_{\text{exp}}, \beta_{\text{pose}}) = \hat{\xbf}_{\text{FLAME}(\mathbf{0}, \mathbf{0})} - \hat{\xbf}_{\text{FLAME}(\beta_{\text{exp}}, \beta_{\text{pose}})}
\end{smequation}
where, \(\hat{\xbf}_{\text{FLAME}(\mathbf{0}, \mathbf{0})}\) is the position of \(\xbf\) in the canonical space and \(\hat{\xbf}_{\text{FLAME}(\beta_{\text{exp}}, \beta_{\text{pose}})}\) is its position with head pose and facial expression parameters \(\{\beta_{\text{exp}}, \beta_{\text{pose}}\}\).

The RigNeRF deformation field can now be defined as the sum of the 3DMM deformation field and the residual predicted by \(D\), as follows
\begin{smequation}
    \begin{split}
        \hat{D}(\xbf) &= \text{3DMMDef}(\xbf, \beta_{\text{i,exp}}, \beta_{\text{i,pose}}) \\
                       & + D(\gammabf_{a}(\xbf), \gammabf_{b}(\text{3DMMDef}(\xbf, \beta_{\text{i,exp}}, \beta_{\text{i,pose}})), \omega_{i})\\
        \xbfcan =  & \xbf +  \hat{D}(\xbf)
    \end{split}
    \label{eq:rignerfcanmap}
\end{smequation}
where, \(\hat{D}(\xbf)\) is the value of the RigNeRF deformation field at \(\xbf\), \(\{\gammabf_{a}, \gammabf_{b}\}\) is the positional embedding on \(\xbf\) and \(\text{3DMMDef}(\xbf,...)\) respectively and \(\omega_{i}\) is the deformation embedding for current frame. We use \(\omega_{i}\) to model deformations that cannot be accounted for by head-pose and expression changes. 
Experimentally, we find that conditioning \(D\) directly on the expression and pose parameters,  \(\{\beta_{\text{i,exp}}, \beta_{\text{i,pose}}\}\), leads to overfitting and poor generalization. This is likely due to the high dimensionality  of the code (100), that makes it prone to overfitting. Instead, we condition \(D\) on the 3DMM deformation of the point \(\xbf\), \(\text{3DMMDef}(\xbf, \beta_{\text{i,exp}}, \beta_{\text{i,pose}})\). Since \(\text{3DMMDef}(\xbf, \beta_{\text{i,exp}}, \beta_{\text{i,pose}}) \in \mathbb{R}^{3}\), it is itself relatively low dimensional, and it can be pushed into higher dimensions by adjusting the number of frequencies of its positional embedding, \(\gammabf_{b}\). We find that using \(b = 2\) frequencies in \(\gammabf_{b}\) for the 3DMM deformation, \(\text{3DMMDef}(\xbf, \beta_{\text{i,exp}}, \beta_{\text{i,pose}})\), works the best. Additionally, we transform the all the points by the inverse of head-pose rotation. We do this so that the deformation networks acts on 3D points that are in a space where the head is roughly aligned across all frames.\footnote{We thank Reviewer 1 of CVPR 2022 for a similar suggestion.}

\begin{figure}[h]
    \centering
    \includegraphics[width=0.9\linewidth]{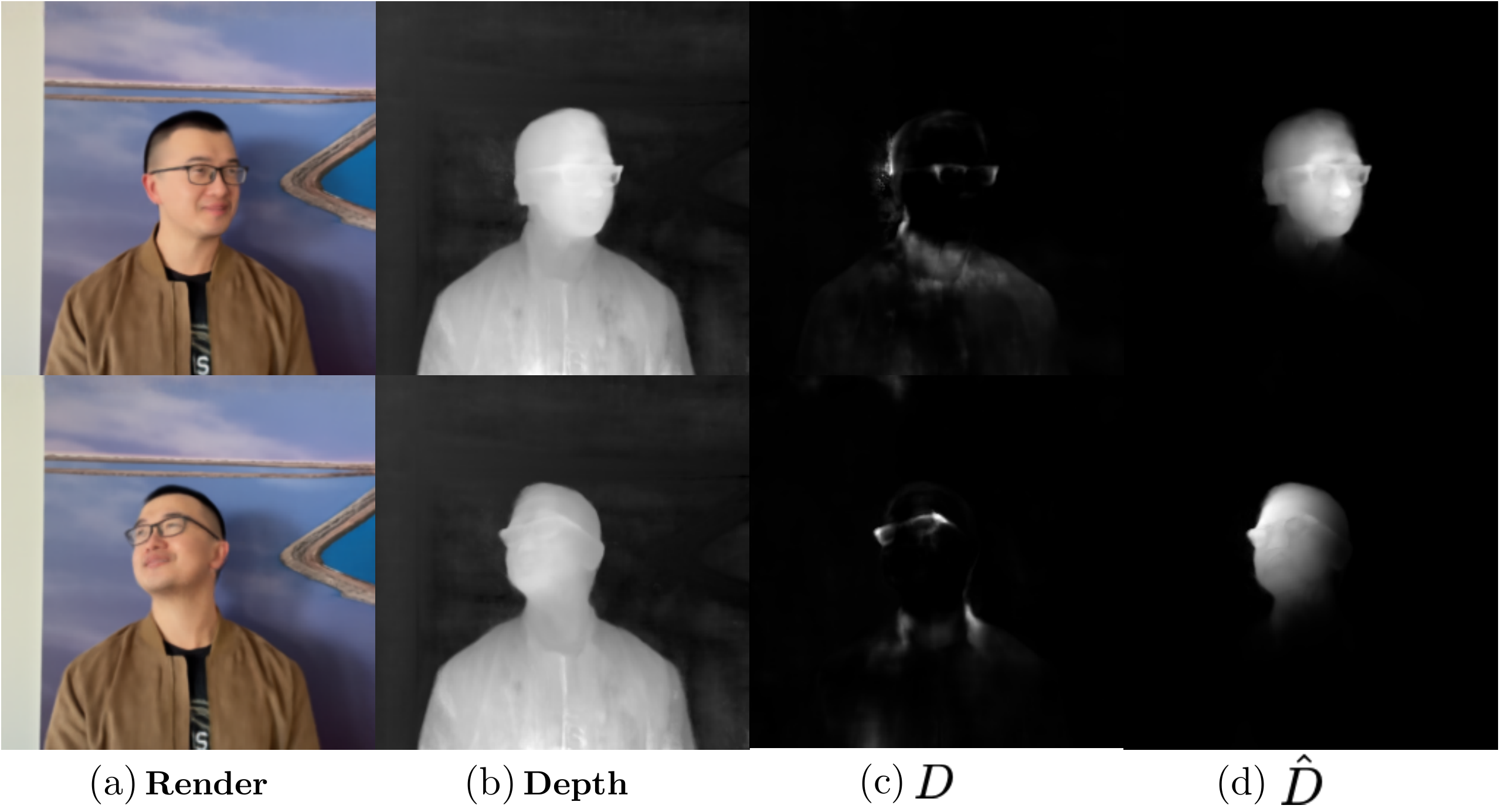}
    
    \caption{{\textbf{Visualize learnt depth and deformation in \MethodName.} Here we show the depth, the magnitude of the output of the Deformation MLP \(D\), and magnitude of the \(\hat{D}\) i.e the sum of the 3DMM Deformation and \(D\). In (b), we can see that despite large changes in head-pose, the depth remains consistent. Next, in (c), we see \(D\) generates a deformation around the glasses for both poses so that it can be accurately deformed along with the head. Finally, in the last column we see how \(\hat{D}\) is only concentrated on the head.
    }}
    \vspace{-3mm}
    \label{fig:depth-view}
\end{figure}

In \fig{depth-view}, we show renders of both the output of \(D\) and the  RigNeRF deformation field, \(\hat{D}\), as described in \eq{rignerfcanmap}. In \fig{depth-view}(c), we see that \(D\) generates the accurate deformation around the glasses, which the 3DMM deformation cannot do, for both head-poses. In \fig{depth-view}(d), we see that the \(\hat{D}\) is only concentrated on the head as it should be.

\subsection{3DMM-conditioned Appearance}

In order to accurately model expression and head-pose based textures, such as teeth, we condition \(F\) on both expression and head-pose parameters and on features  extracted from the penultimate layer of the deformation MLP \(D(\gammabf_{a}(\xbf),...)\). We find that using these features as input improves the overall quality of the render, please check the supplementary for details. Thus, once a point \(\xbf\) has been deformed to its location in the canonical space, \(\xbfcan\), using \eq{rignerfcanmap}, its color is calculated as follows:
\begin{smequation}
        \mathbf{c}(\xbf, \mathbf{d}), \sigma(\xbf) = F(\gammabf_{c}(\xbfcan), \gammabf_{d}(\mathbf{d}), \phi_{i}, D_{F,i}(\xbfcan), \beta_{\text{i,exp}}, \beta_{\text{i,pose}})
        \label{eq:rig_nerf}
\end{smequation}
where, \(\mathbf{d}\) is the viewing direction, \(\gammabf_{c}, \gammabf_{d}\) is the positional embedding on \(\xbfcan\) and \(\mathbf{d}\), and \(D_{F,i}(\xbfcan)\) are features from the penultimate layer of the deformation MLP \(D(\gammabf_{a}(\xbf),...)\). The pixel color for \(p\) is then calculated using volume rendering and parameters of \MethodName are minimized w.r.t to the ground truth color of \(p\). The full architecture is shown in \fig{method}.
\section{Results}

\begin{figure*}[h]
    \centering
    \includegraphics[width=\linewidth]{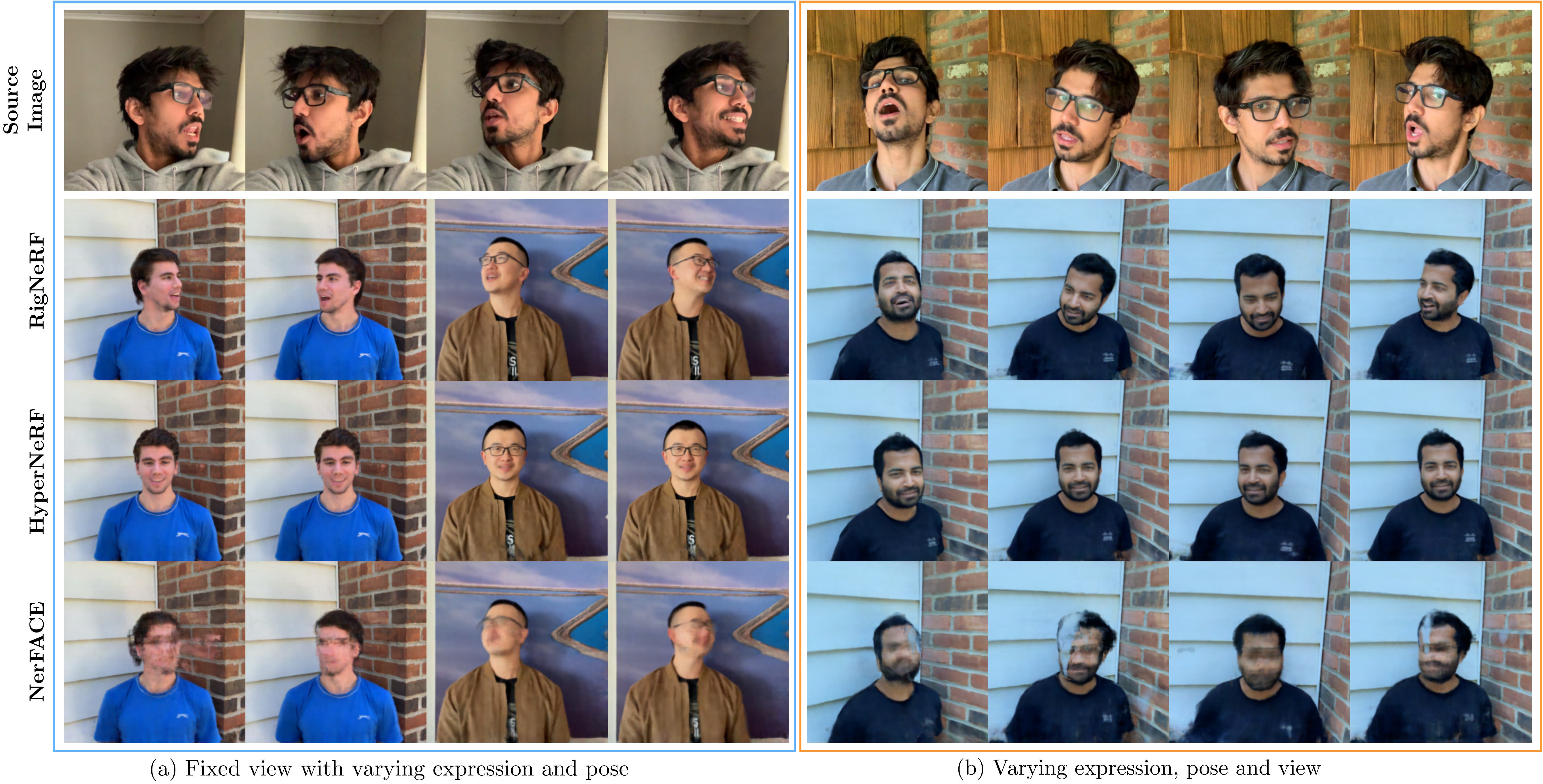}
    
    \caption{{\textbf{Qualitative comparison by reanimation with novel facial expression, head-pose, and camera view parameters.} Here we reanimate \MethodName, HyperNeRF \cite{park2021hypernerf} and NerFACE \cite{gafni2020dynamic} using facial expression and head-pose derived from source images (top-row). We observe that while HyperNeRF~\cite{park2021hypernerf} is able to generate realistic looking images of a portrait, it is unable to control head pose or facial expression in the result. %
    On the other hand, NerFACE\cite{gafni2020dynamic} attempts to render the correct pose and expression, but is unable to generate plausible face regions. Since, NerFACE \cite{gafni2020dynamic} lacks an explicit deformation module it is unable to model deformation due to head-pose and facial expression changes. 
    In contrast, our approach \MethodName can effectively control the head pose, facial expression, and camera view, generating high quality facial appearance. 
    }}
    \label{fig:view_exp}
\end{figure*}

\begin{table*}[t]
\begin{center}
\small
\scalebox{0.75}{
\begin{tabular}{lccccccccccccc}
\toprule
  & \multicolumn{3}{c}{\emph{Subject 1}}  &  \multicolumn{3}{c}{\emph{Subject 2}} &  \multicolumn{3}{c}{\emph{Subject 3}} &  \multicolumn{3}{c}{\emph{Subject 4}}\\
  \midrule
  Models   & PSNR $\uparrow$ &LPIPS $\downarrow$ & FaceMSE $\downarrow$  
   & PSNR $\uparrow$ &LPIPS $\downarrow$ & FaceMSE $\downarrow$
  & PSNR $\uparrow$ &LPIPS $\downarrow$ & FaceMSE $\downarrow$
  & PSNR $\uparrow$ &LPIPS $\downarrow$ & FaceMSE $\downarrow$\\
\midrule
 \MethodName (Ours) &
 \hl{29.55} & \hl{0.136} & \hl{9.6e-5} &
 \hl{29.36} & \hl{0.102} & \hl{1e-4} &
 \hl{28.39} & \hl{0.109} & \hl{8e-5} &
 27.0 & \hl{0.092} & \hl{2.3e-4}
 \\
 HyperNeRF \cite{park2021hypernerf} &
 24.58 &  0.22  & 8.14e-4  & 
 22.55 &  0.1546  & 9.48e-4 &
 19.29 &  0.26  & 2.74e-3 &
 21.19 &  0.182  & 1.58e-3
  \\
 NerFACE \cite{NerFACE} &
  24.2 &  0.217  & 7.84e-4  & 
  24.57 &  0.174  & 6.7e-4 &
  28.00 &  0.1292  & 1.2e-4 &
  \hl{28.47} &  0.134  & 2.7e-4
 \\ 
 FOMM \cite{siarohin2020first} &
  11.45 &  0.432  & 7.65e-3  & 
  12.7 &  0.582  & 6.31e-3 &
  10.17 &  0.601  & 1.7e-2 &
  11.17 &  0.529  & 6.8e-3
 \\ 
\bottomrule
\end{tabular}}
\caption{Quantitative results of Subject 1,2,3 and 4 on test data. Our results are better than HyperNeRF \cite{park2021hypernerf}, NerFACE \cite{gafni2020dynamic} and FOMM \cite{siarohin2020first} on most metrics across all subjects. 
}
\vspace{-1.0cm}
\label{tab:Subjects_metrics}
\end{center}
\end{table*}

\begin{figure*}[h]
    \centering
    \includegraphics[width=0.95\linewidth]{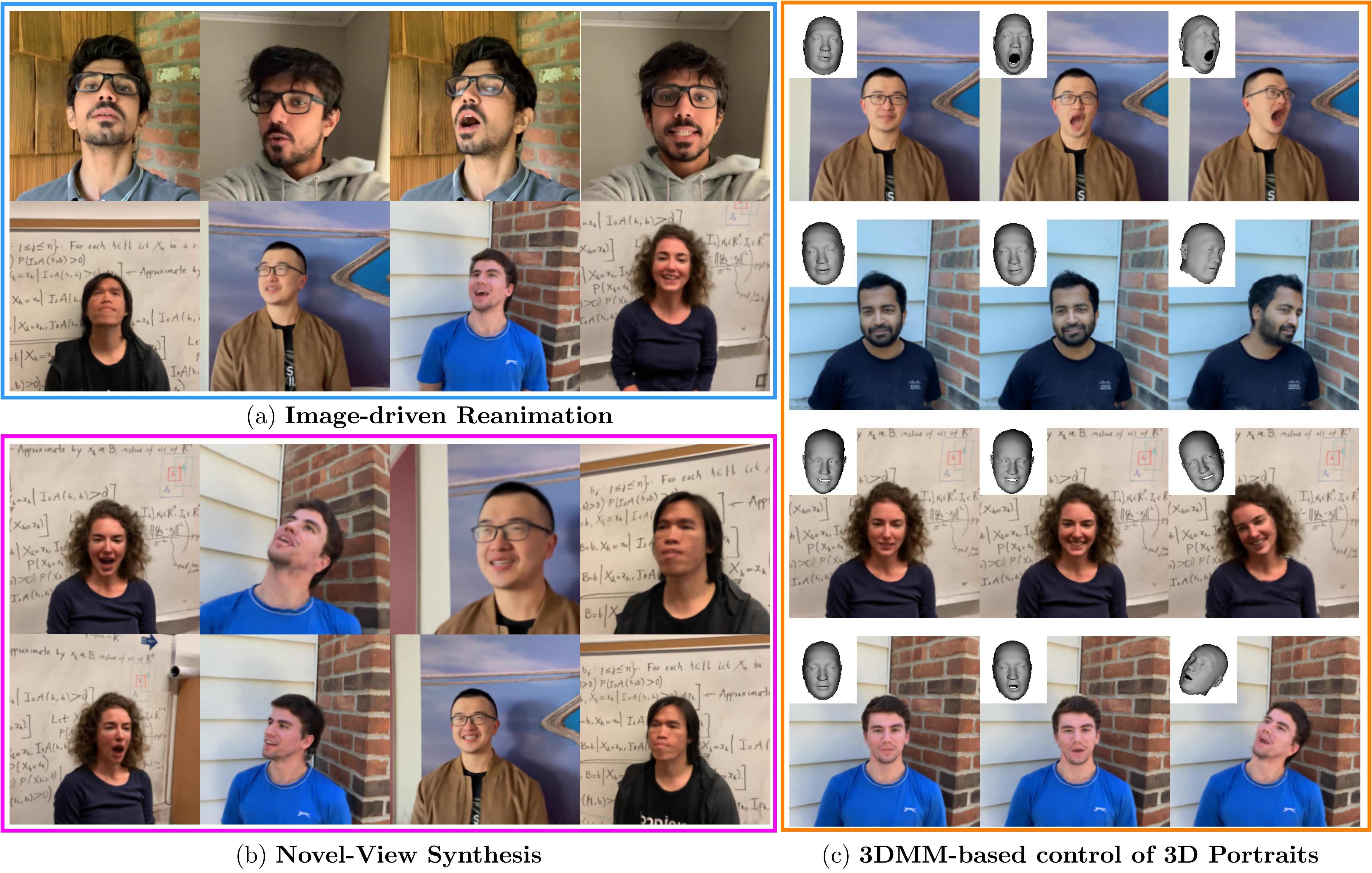}
    
    \caption{{Applications of \textbf{\MethodName}. \MethodName allows for full control head-pose, facial expressions and viewing direction of 3D portrait scenes. This enables application like (a) Image-driven Reanimation, (b) Novel View Synthesis and (c) 3DMM-based control of 3D Portraits. In (a)-top row, we show images of ``driving sequences'' from which we extract pose and expression parameters; the results from 4 subjects are shown in (a)-bottom where we synthesize realistic portrait frames that closely matches the driving pose and expression. We show in (b) a set of view synthesis results where we fix the head pose and facial expression, rendering results with varying camera positions, in which we show high-quality results with dramatic view changes. In (c), we demonstrate the application of controlling the portrait appearance with explicit 3DMM parameters. Within each row, (c)-column 1 and (c)-column 2 have the same pose but different expressions; (c)-column 2 and (c)-column 3 have the different poses but share the same expression. The inset shows the input 3DMM pose and expression, both of which are faithfully rendered in the corresponding results. Please find more results in the supplemental document and video.
    }}
    \vspace{-0.5cm}
    \label{fig:all_control}
\end{figure*}

\begin{figure}[h]
    \centering
    \includegraphics[width=0.95\linewidth]{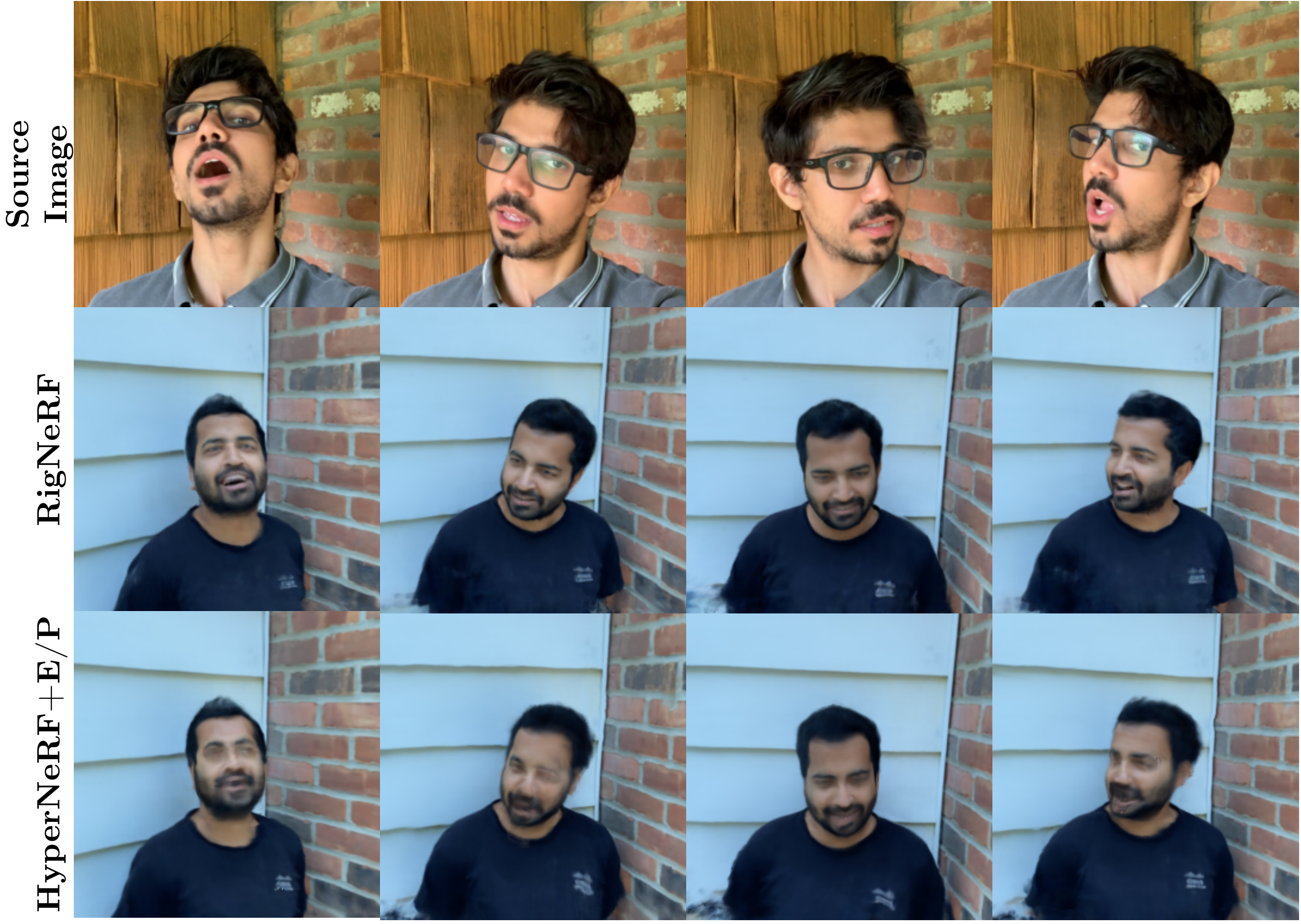}
    
    \caption{{\textbf{A qualitative comparison between \MethodName and HyperNeRF+E/P.} Here we show a qualitative comparison between \MethodName and HyperNeRF+E/P when reanimated using source images. We see that HyperNeRF+E/P generates many artefacts during reanimation due to its inability to model deformations correctly. In constrast, \MethodName generates realistic reanimations with high fidelity to both the head-pose and facial expressions.
    }}
    \vspace{-0.5cm}
    \label{fig:Hypernerfpp_comp}
\end{figure}

In this section, we show results of head-pose control, facial expression control, and novel view synthesis using \MethodName. For each scene, the model is trained on a short portrait video captured using a consumer smartphone. 

\paragraph{Baseline approaches} To the best of our knowledge, \MethodName is the first method that enables dynamic control of head-pose, facial expressions along with the ability to synthesize novel views of full portrait scenes. Thus, there is no existing work for an apple-to-apple comparison. We qualitatively and quantitatively compare our method to three other methods that perform closely related tasks: (1) HyperNeRF \cite{park2021hypernerf}: a state-of-the-art method using NeRFs for novel view synthesis of dynamic portrait scenes, \textit{without} any control (2) NerFACE \cite{gafni2020dynamic}, a state-of-the-art method using NeRF for face dynamics control \textit{without} modeling camera viewpoint and the entire scene, and (3) First Order Motion Model (FOMM) \cite{siarohin2020first}, a general-purpose image reanimation pipeline. 

When generating renanimation videos, \MethodName, HyperNeRF \cite{park2021hypernerf} and NerFACE \cite{gafni2020dynamic} require an appearance code for rendering; we use the appearance code of the first frame here. Similarly, \MethodName and Nerfies \cite{nerfies} require a deformation code and we use the deformation code from the first frame. Full videos of the reanimation can be found in the supplementary material. We strongly urge the readers to refer to the videos to evaluate the quality of the results. %

\paragraph{Training Data Capture and Training details} The training and validation data was captured using an iPhone XR or iPhone 12 for all the experiments in the paper. In the first half of the capture, we ask the subject to enact a wide range of expressions and speech while trying to keep their head still as the camera is panned around them. In the next half, the camera is fixed at head-level and the subject is asked to rotate their head as they enact a wide range of expressions. Camera parameters are calculated using COLMAP \cite{Schonberger-2016-SFM}. We calculate the expression and shape parameters of each frame in the videos using DECA \cite{DECA} and further optimize them using via standard landmark fitting using the landmarks predicted by \cite{3DDFA_V2} and camera parameters given by COLMAP \cite{Schonberger-2016-SFM}. All training videos are between 40-70 seconds long (\(\sim\) 1200-2100 frames)%
. Due to compute restrictions, the video is down-sampled and the models are trained at 256x256 resolution. We use coarse-to-fine and vertex deformation regularization \cite{nerfies} to train the deformation network \(D(\xbf, \omega_{i})\). Please find full details of each experiment in the supplementary.
\vspace{-0.1cm}
\subsection{Evaluation on Test Data}
\label{sect:eval_data}
We evaluate \MethodName, HyperNeRF \cite{park2021hypernerf}, NerFACE \cite{gafni2020dynamic} and FOMM \cite{siarohin2020first} on held out images on the captured video sequences. We use the camera view, pose and expression parameters of these images.
Since \MethodName and HyperNeRF \cite{park2021hypernerf}  use a per-frame deformation \(\omega_{i}\), we can't use the first frame (which is what we use as default for reanimation) to perform a direct comparison with the ground truth image as it may have a different deformation to the canonical space than the first frame. Therefore, we first optimize for the deformation code, \(\omega_{v}\), of a given validation image by minimizing rendering error wrt that frame as follows:
\begin{equation}
    \omega_{v} = \underset{\omega}{\text{min}} || C_{p}(\omega; {\xbf}, \mathbf{d}, \theta, \phi_{0}, \beta_{\text{i,exp}}, \beta_{\text{i,pose}}) -  C_{p}^{GT} ||
    \label{eq:def_opt}
\end{equation}
where, \(C_{p}(\omega; {\xbf}, \mathbf{d}, \theta, \phi_{0})\) is the predicted color at pixel \(p\) generated using \eq{rig_nerf} and volume-rendering , \(\phi_{0}\) is the appearance code of the first frame, \(\theta\) are the parameters of \(F\) as defined in \eq{rig_nerf} and \(C_{p}^{GT}\) is the ground-truth pixel value. Note, we \textit{only} optimize \(\omega\), all other parameters of the radiance field are kept fixed. We optimize \eq{def_opt} for 200 epochs which we observe to be more than enough to find the loss plateau. Once the optimization finishes, we report the final MSE, PSNR, LPIPS and Face MSE i.e the MSE only over the face region. 
We use no such optimization with NerFACE \cite{gafni2020dynamic} since it does not have a deformation module and on FOMM \cite{siarohin2020first} since it is a image-based method. 

As can be seen in \tab{Subjects_metrics}, our method outperforms HyperNeRF \cite{park2021hypernerf}, NerFACE \cite{gafni2020dynamic} and FOMM \cite{siarohin2020first} on held-out test images. \MethodName, HyperNeRF \cite{park2021hypernerf} and NerFACE \cite{gafni2020dynamic} are trained on dynamic portrait videos with changing head-pose and facial expressions, HyperNeRF \cite{park2021hypernerf}, lacking any head-pose and facial expression control, is unable to generate the head-poses and facial expressions seen in the held-out test set. Even as we optimized the deformation code of HyperNeRF \cite{park2021hypernerf}, we found it hard to fit the unseen test images.  NerFACE \cite{gafni2020dynamic}, on the other hand, lacking a deformation module, therefore is unable to model the dynamism of head-pose changes solely by concatenating pose and expression parameters as input the NeRF MLP. As a result, NerFACE \cite{gafni2020dynamic} generates significant artefacts on the face regions (see the third row of \fig{view_exp}(a) and \fig{view_exp}(b)). FOMM \cite{siarohin2020first}, being an image based method, is unable to model novel views. Qualitative results of FOMM \cite{siarohin2020first} can be found in the supplementary. In contrast to other methods, \MethodName, thanks to the use of a 3DMM-guided deformation module, is able to model head-pose, facial expressions and the full 3D portrait scene with high fidelity, thus giving better reconstructions with sharp details.

\subsection{Reanimation with pose and expression control}
\label{sect:reanim}
In this section we show results of reanimating a portrait video using both \MethodName, HyperNeRF \cite{park2021hypernerf} and NerFACE \cite{gafni2020dynamic} using expression and head-pose parameters as the driving parameters. Per-frame expression and head-pose parameters from the driving video are extracted using DECA \cite{DECA}+Landmark fitting \cite{3DDFA_V2} and are given as input to \MethodName in \eq{rignerfcanmap} and \eq{rig_nerf}.
 Since HyperNeRF \cite{park2021hypernerf} does not take as input head-pose or expression parameters, it's forward pass remains unchanged.

 First, in \fig{view_exp}(a) we show the results of changing head-pose and expression using a driving video while keeping view constant. As one can see, \MethodName captures the driving head-pose and facial expressions with high fidelity without compromising the reconstruction of the entire 3D scene. In contrast, we see that HyperNeRF \cite{park2021hypernerf} is unable to change facial expressions or head-pose due to the lack of controls, while NerFACE \cite{gafni2020dynamic} generates significant artefacts on the face, especially when head-pose is changed. In \fig{view_exp}(b), we show the results of \textit{novel view synthesis along with changing head-pose and facial expressions}. Again, we see that \MethodName reanimates the subject with the accurate head-pose and facial expressions \textit{and is able to do so regardless of viewing direction} and without compromising the reconstruction of the background 3D scene. Again, HyperNeRF is unable to change the  head-pose and facial expressions due to the aforementioned reason while NerFACE \cite{gafni2020dynamic} generates significant artefacts during reanimation. 
 
 In \fig{all_control}, we show more qualitative results of \MethodName. We use three different applications of \MethodName to demonstrate its flexibility and full controllability of a portrait scene. In \fig{all_control}-(a), we show additional results of image-driven animation. The result frames (\fig{all_control}-(a)-bottom) closely reproduce the head pose and facial expression shown in the driving frames (\fig{all_control}-(a)-top). In \fig{all_control}-(b), we show results of varying camera views while fixing (an arbitrary) facial expression and head pose. We demonstrate robust view synthesis performance with dramatic view changes. In \fig{all_control}-(c), we show that \MethodName can take user-specified 3DMM parameters as input to generate high-quality portrait images: each frame faithfully reproduces the face and expression provided by a set of 3DMM parameters shown in the inset.

\vspace{-0.1cm}
\subsection{Comparison with HyperNeRF+E/P}
\label{sect:hypernerfpp}
In this section we compare against HyperNeRF \cite{park2021hypernerf} with added pose and expression control, which we named HyperNeRF+E/P. The forward pass of this model is as follows
\begin{smequation}
\begin{split}
    \xbfcan &= D(\gammabf_{a}(\xbf),  \beta_{\text{i,exp}}, \beta_{\text{i,pose}}, \omega_{i})\\
    \textbf{w} &= H(\gammabf_{l}(\xbf), \omega_{i})\\
    \mathbf{c}(\xbf, \mathbf{d}), \sigma(\xbf) &= F(\gammabf_{c}(\xbfcan), \gammabf_{d}(\mathbf{d}), \phi_{i}, \textbf{w} , \beta_{\text{i,exp}},\beta_{\text{i,pose}})
\end{split}
\end{smequation}
where, \(H\) is the ambient MLP and \(\textbf{w}\) are the ambient coordinates \cite{park2021hypernerf}. 
In \tab{aba_metrics}, we show a quantitative comparison between \MethodName and HyperNeRF+E/P. We see that \MethodName is able to generate better reconstructions of the face and generates images that are perceptually closer to the ground truth, as measured by LPIPS \cite{zhang2018perceptual}, than those generated by HyperNeRF+E/P. In \fig{Hypernerfpp_comp} we show a qualitative comparison when both methods are reanimated using source images, where the head-pose and expression significantly differ from the training set. As can be seen, while HyperNeRF+E/P is able to turn the head, it is unable to do so rigidly (see row 3, column 2 and row 3, column 4 of \fig{Hypernerfpp_comp}). Further, it is also unable to model facial expressions accurately and generates artefacts on the face region. This further demonstrates the benefit of our 3DMM-guided deformation learning.
\begin{table}[t]
\begin{center}
\small
\scalebox{0.7}{
\begin{tabular}{lccccccc}
\toprule
  & \multicolumn{3}{c}{\emph{Subject 1}}  &  \multicolumn{3}{c}{\emph{Subject 2}} \\
  \midrule
  Models   & PSNR $\uparrow$ &LPIPS $\downarrow$ & FaceMSE $\downarrow$  
   & PSNR $\uparrow$ &LPIPS $\downarrow$ & FaceMSE $\downarrow$\\
\midrule
 \MethodName (Ours) &
 29.55 & \hl{0.136} & \hl{9.6e-5} &
 29.36 & \hl{0.102} & \hl{1e-4} 
 \\
 HyperNeRF+Exp  &
 \hl{31.3} &  0.161  & 1.3e-4  & 
 \hl{30} &  0.116  & 1.9e-4 
 \\
  
\bottomrule
\end{tabular}}
\vspace{3pt}
\caption{\small{Quantitative comparison between \MethodName and HyperNeRF+E/P. We see that \MethodName generates better reconstruction of the face with lower perceptual distance to the ground-truth as compared to HyperNeRF+E/P.
}}
\vspace{-1cm}
\label{tab:aba_metrics}
\end{center}
\end{table}

\vspace{-0.3cm}
\section{Limitations and Conclusion}

Our method has certain limitations. First, it is subject specific and trains an individual model for each scene. Due to the need to capture sufficient expression and head-pose variations for training, out method currently requires a training sequences ranging from 40-70 seconds. Additionally, like all other NeRF-based methods, the quality of camera-registration affects the quality of the results. 
Being a method that allows photorealistic facial reanimation, \MethodName may have potentially negative societal impact if misused. We discuss this further in the supplementary.

In conclusion, we present \MethodName, a volumetric neural rendering model for fully controllable human portraits. Once trained, it allows full control of head-pose, facial expression, and viewing direction. In order to ensure generalization to novel head-pose and facial expression we use 3DMM-guided deformation field.
This deformation field allows us to effectively model and control both the rigid deformations caused by head-pose change and non-rigid deformations of changes in facial expressions. Training with a short portrait video, \MethodName enables applications includes image-based face reanimation, portrait novel-view synthesis, and 3DMM-based control of 3D portraits.

\vspace{5pt}
\noindent\textbf{Acknowledgements}\\
\small{We would like to thank the anonymous CVPR reviewers for taking the time to review and suggest improvements to the paper. ShahRukh Athar is supported by a gift from Adobe, Partner University Fund 4DVision Project, and the
SUNY2020 Infrastructure Transportation Security Center.}

{
    \clearpage
    \small
    \bibliographystyle{ieee_fullname}
    \bibliography{macros,main}

\begin{thebibliography}{10}\itemsep=-1pt

\bibitem{facedet3d}
ShahRukh Athar, Albert Pumarola, Francesc Moreno-Noguer, and Dimitris Samaras.
\newblock Facedet3d: Facial expressions with 3d geometric detail prediction.
\newblock {\em arXiv preprint arXiv:2012.07999}, 2020.

\bibitem{athar2020self}
S Athar, Z Shu, and D Samaras.
\newblock Self-supervised deformation modeling for facial expression editing.
\newblock 2020.

\bibitem{Bemana-2020-XIN}
Mojtaba Bemana, Karol Myszkowski, Hans-Peter Seidel, and Tobias Ritschel.
\newblock X-fields: Implicit neural view-, light-and time-image interpolation.
\newblock 2020.

\bibitem{blanz1999morphable}
Volker Blanz, Thomas Vetter, et~al.
\newblock A morphable model for the synthesis of 3d faces.
\newblock 1999.

\bibitem{chatziagapi2021sider}
Aggelina Chatziagapi, ShahRukh Athar, Francesc Moreno-Noguer, and Dimitris
  Samaras.
\newblock Sider: Single-image neural optimization for facial geometric detail
  recovery.
\newblock {\em arXiv preprint arXiv:2108.05465}, 2021.

\bibitem{SRF}
Julian Chibane, Aayush Bansal, Verica Lazova, and Gerard Pons-Moll.
\newblock Stereo radiance fields (srf): Learning view synthesis for sparse
  views of novel scenes.
\newblock In {\em CVPR}, 2021.

\bibitem{StarGAN2018}
Yunjey Choi, Minje Choi, Munyoung Kim, Jung-Woo Ha, Sunghun Kim, and Jaegul
  Choo.
\newblock Stargan: Unified generative adversarial networks for multi-domain
  image-to-image translation.
\newblock In {\em CVPR}, 2018.

\bibitem{starganv2}
Yunjey Choi, Youngjung Uh, Jaejun Yoo, and Jung-Woo Ha.
\newblock Stargan v2: Diverse image synthesis for multiple domains.
\newblock In {\em CVPR}, 2020.

\bibitem{deng2020disentangled}
Yu Deng, Jiaolong Yang, Dong Chen, Fang Wen, and Xin Tong.
\newblock Disentangled and controllable face image generation via 3d
  imitative-contrastive learning.
\newblock In {\em Proceedings of the IEEE/CVF Conference on Computer Vision and
  Pattern Recognition}, pages 5154--5163, 2020.

\bibitem{Doukas2021Head2HeadDF}
M. Doukas, Mohammad~Rami Koujan, V. Sharmanska, A. Roussos, and S. Zafeiriou.
\newblock Head2head++: Deep facial attributes re-targeting.
\newblock {\em IEEE Transactions on Biometrics, Behavior, and Identity
  Science}, 3:31--43, 2021.

\bibitem{DECA}
Yao Feng, Haiwen Feng, Michael~J. Black, and Timo Bolkart.
\newblock Learning an animatable detailed {3D} face model from in-the-wild
  images.
\newblock volume~40, 2021.

\bibitem{gafni2020dynamic}
Guy Gafni, Justus Thies, Michael Zollh{\"o}fer, and Matthias Nie{\ss}ner.
\newblock Dynamic neural radiance fields for monocular 4d facial avatar
  reconstruction.
\newblock In {\em CVPR}, June 2021.

\bibitem{NerFACE}
Guy Gafni, Justus Thies, Michael Zollhöfer, and Matthias Nießner.
\newblock Dynamic neural radiance fields for monocular 4d facial avatar
  reconstruction, 2020.

\bibitem{Gao-portraitnerf}
Chen Gao, Yichang Shih, Wei-Sheng Lai, Chia-Kai Liang, and Jia-Bin Huang.
\newblock Portrait neural radiance fields from a single image.
\newblock {\em arXiv preprint arXiv:2012.05903}, 2020.

\bibitem{goodfellow2014generative}
Ian Goodfellow, Jean Pouget-Abadie, Mehdi Mirza, Bing Xu, David Warde-Farley,
  Sherjil Ozair, Aaron Courville, and Yoshua Bengio.
\newblock Generative adversarial nets.
\newblock In {\em NeurIPS}, 2014.

\bibitem{3DDFA_V2}
Jianzhu Guo, Xiangyu Zhu, Yang Yang, Fan Yang, Zhen Lei, and Stan~Z Li.
\newblock Towards fast, accurate and stable 3d dense face alignment.
\newblock In {\em Proceedings of the European Conference on Computer Vision
  (ECCV)}, 2020.

\bibitem{pix2pix2016}
Phillip Isola, Jun-Yan Zhu, Tinghui Zhou, and Alexei~A Efros.
\newblock Image-to-image translation with conditional adversarial networks.
\newblock In {\em CVPR}, 2017.

\bibitem{StyleGAN}
Tero Karras, Samuli Laine, and Timo Aila.
\newblock A style-based generator architecture for generative adversarial
  networks.
\newblock In {\em CVPR}, 2019.

\bibitem{Karras-2019-ASB}
Tero Karras, Samuli Laine, and Timo Aila.
\newblock A style-based generator architecture for generative adversarial
  networks.
\newblock In {\em CVPR}, 2019.

\bibitem{Karras-2020-AAI}
Tero Karras, Samuli Laine, Miika Aittala, Janne Hellsten, Jaakko Lehtinen, and
  Timo Aila.
\newblock Analyzing and improving the image quality of stylegan.
\newblock In {\em CVPR}, 2020.

\bibitem{Kim2018DeepVP}
H. Kim, P. Garrido, A. Tewari, Weipeng Xu, Justus Thies, M. Nie{\ss}ner,
  Patrick P{\'e}rez, C. Richardt, M. Zollh{\"o}fer, and C. Theobalt.
\newblock Deep video portraits.
\newblock {\em ACM Transactions on Graphics (TOG)}, 37:1 -- 14, 2018.

\bibitem{Kim-2018-DVP}
Hyeongwoo Kim, Pablo Garrido, Ayush Tewari, Weipeng Xu, Justus Thies, Matthias
  Niessner, Patrick P{\'e}rez, Christian Richardt, Michael Zollh{\"o}fer, and
  Christian Theobalt.
\newblock Deep video portraits.
\newblock {\em ACM TOG}, 2018.

\bibitem{head2head2020}
M. Koujan, M. Doukas, A. Roussos, and S. Zafeiriou.
\newblock Head2head: Video-based neural head synthesis.
\newblock In {\em 2020 15th IEEE International Conference on Automatic Face and
  Gesture Recognition (FG 2020) (FG)}, pages 319--326, Los Alamitos, CA, USA,
  may 2020. IEEE Computer Society.

\bibitem{kowalski2020config}
Marek Kowalski, Stephan~J Garbin, Virginia Estellers, Tadas Baltru{\v{s}}aitis,
  Matthew Johnson, and Jamie Shotton.
\newblock Config: Controllable neural face image generation.
\newblock In {\em Computer Vision--ECCV 2020: 16th European Conference,
  Glasgow, UK, August 23--28, 2020, Proceedings, Part XI 16}, pages 299--315.
  Springer, 2020.

\bibitem{lassner2020pulsar}
Christoph Lassner and Michael ZollhÃ¶fer.
\newblock Pulsar: Efficient sphere-based neural rendering.
\newblock In {\em CVPR}, 2021.

\bibitem{FLAME:SiggraphAsia2017}
Tianye Li, Timo Bolkart, Michael.~J. Black, Hao Li, and Javier Romero.
\newblock Learning a model of facial shape and expression from {4D} scans.
\newblock {\em ACM Transactions on Graphics, (Proc. SIGGRAPH Asia)}, 36(6),
  2017.

\bibitem{li2021neural}
Tianye Li, Mira Slavcheva, Michael Zollhoefer, Simon Green, Christoph Lassner,
  Changil Kim, Tanner Schmidt, Steven Lovegrove, Michael Goesele, and Zhaoyang
  Lv.
\newblock Neural 3d video synthesis, 2021.

\bibitem{NeuralSceneFlow}
Zhengqi Li, Simon Niklaus, Noah Snavely, and Oliver Wang.
\newblock Neural scene flow fields for space-time view synthesis of dynamic
  scenes.
\newblock In {\em CVPR}, 2021.

\bibitem{Liu-2020-NSV}
Lingjie Liu, Jiatao Gu, Kyaw~Zaw Lin, Tat-Seng Chua, and Christian Theobalt.
\newblock Neural sparse voxel fields.
\newblock 2020.

\bibitem{liu2021neural}
Lingjie Liu, Marc Habermann, Viktor Rudnev, Kripasindhu Sarkar, Jiatao Gu, and
  Christian Theobalt.
\newblock Neural actor: Neural free-view synthesis of human actors with pose
  control.
\newblock {\em ACM TOG}, 2021.

\bibitem{lombardi2021mixture}
Stephen Lombardi, Tomas Simon, Gabriel Schwartz, Michael Zollhoefer, Yaser
  Sheikh, and Jason Saragih.
\newblock Mixture of volumetric primitives for efficient neural rendering,
  2021.

\bibitem{Martin-2020-NIT}
Ricardo Martin-Brualla, Noha Radwan, Mehdi~SM Sajjadi, Jonathan~T Barron,
  Alexey Dosovitskiy, and Daniel Duckworth.
\newblock Nerf in the wild: Neural radiance fields for unconstrained photo
  collections.
\newblock {\em arXiv:2008.02268}, 2020.

\bibitem{nerf}
Ben Mildenhall, Pratul~P Srinivasan, Matthew Tancik, Jonathan~T Barron, Ravi
  Ramamoorthi, and Ren Ng.
\newblock Nerf: Representing scenes as neural radiance fields for view
  synthesis.
\newblock 2020.

\bibitem{unisurf}
Michael Oechsle, Songyou Peng, and Andreas Geiger.
\newblock Unisurf: Unifying neural implicit surfaces and radiance fields for
  multi-view reconstruction.
\newblock {\em arXiv preprint arXiv:2104.10078}, 2021.

\bibitem{nerfies}
Keunhong Park, Utkarsh Sinha, Jonathan~T. Barron, Sofien Bouaziz, Dan~B
  Goldman, Steven~M. Seitz, and Ricardo Martin-Brualla.
\newblock Nerfies: Deformable neural radiance fields.
\newblock {\em ICCV}, 2021.

\bibitem{park2021hypernerf}
Keunhong Park, Utkarsh Sinha, Peter Hedman, Jonathan~T. Barron, Sofien Bouaziz,
  Dan~B Goldman, Ricardo Martin-Brualla, and Steven~M. Seitz.
\newblock Hypernerf: A higher-dimensional representation for topologically
  varying neural radiance fields.
\newblock {\em arXiv preprint arXiv:2106.13228}, 2021.

\bibitem{peng2021animatable}
Sida Peng, Junting Dong, Qianqian Wang, Shangzhan Zhang, Qing Shuai, Xiaowei
  Zhou, and Hujun Bao.
\newblock Animatable neural radiance fields for modeling dynamic human bodies.
\newblock In {\em ICCV}, 2021.

\bibitem{pumarola2020ganimation}
Albert Pumarola, Antonio Agudo, Aleix~M Martinez, Alberto Sanfeliu, and
  Francesc Moreno-Noguer.
\newblock Ganimation: One-shot anatomically consistent facial animation.
\newblock {\em International Journal of Computer Vision}, 128(3):698--713,
  2020.

\bibitem{DNeRF}
Albert Pumarola, Enric Corona, Gerard Pons-Moll, and Francesc Moreno-Noguer.
\newblock {D-NeRF: Neural Radiance Fields for Dynamic Scenes}.
\newblock In {\em CVPR}, 2021.

\bibitem{SVS}
Gernot Riegler and Vladlen Koltun.
\newblock Stable view synthesis.
\newblock In {\em CVPR}, 2021.

\bibitem{Schonberger-2016-SFM}
Johannes~L Schonberger and Jan-Michael Frahm.
\newblock Structure-from-motion revisited.
\newblock In {\em CVPR}, 2016.

\bibitem{shu2018deforming}
Zhixin Shu, Mihir Sahasrabudhe, Riza Alp~Guler, Dimitris Samaras, Nikos
  Paragios, and Iasonas Kokkinos.
\newblock Deforming autoencoders: Unsupervised disentangling of shape and
  appearance.
\newblock In {\em ECCV}, 2018.

\bibitem{NeuralFace2017}
Z. Shu, E. Yumer, S. Hadap, K. Sunkavalli, E. Shechtman, and D. Samaras.
\newblock Neural face editing with intrinsic image disentangling.
\newblock In {\em CVPR}, 2017.

\bibitem{siarohin2020first}
Aliaksandr Siarohin, St{\'e}phane Lathuili{\`e}re, Sergey Tulyakov, Elisa
  Ricci, and Nicu Sebe.
\newblock First order motion model for image animation.
\newblock 2019.

\bibitem{sitzmann2019scene}
Vincent Sitzmann, Michael Zollh{\"o}fer, and Gordon Wetzstein.
\newblock Scene representation networks: Continuous 3d-structure-aware neural
  scene representations.
\newblock 2019.

\bibitem{tewari2020pie}
Ayush Tewari, Mohamed Elgharib, Florian Bernard, Hans-Peter Seidel, Patrick
  P{\'e}rez, Michael Zollh{\"o}fer, and Christian Theobalt.
\newblock Pie: Portrait image embedding for semantic control.
\newblock {\em ACM Transactions on Graphics (TOG)}, 39(6):1--14, 2020.

\bibitem{tewari2020stylerig}
Ayush Tewari, Mohamed Elgharib, Gaurav Bharaj, Florian Bernard, Hans-Peter
  Seidel, Patrick P{\'e}rez, Michael Z{\"o}llhofer, and Christian Theobalt.
\newblock Stylerig: Rigging stylegan for 3d control over portrait images, cvpr
  2020.
\newblock In {\em {IEEE} Conference on Computer Vision and Pattern Recognition
  (CVPR)}. {IEEE}, june 2020.

\bibitem{dnr}
Justus Thies, M. Zollh{\"o}fer, and M. Nie{\ss}ner.
\newblock Deferred neural rendering.
\newblock {\em ACM Transactions on Graphics (TOG)}, 2019.

\bibitem{wang2019cnngenerated}
Sheng-Yu Wang, Oliver Wang, Richard Zhang, Andrew Owens, and Alexei~A Efros.
\newblock Cnn-generated images are surprisingly easy to spot...for now.
\newblock In {\em CVPR}, 2020.

\bibitem{Wizadwongsa2021NeX}
Suttisak Wizadwongsa, Pakkapon Phongthawee, Jiraphon Yenphraphai, and Supasorn
  Suwajanakorn.
\newblock Nex: Real-time view synthesis with neural basis expansion.
\newblock In {\em CVPR}, 2021.

\bibitem{xian2020space}
Wenqi Xian, Jia-Bin Huang, Johannes Kopf, and Changil Kim.
\newblock Space-time neural irradiance fields for free-viewpoint video.
\newblock In {\em CVPR}, 2021.

\bibitem{IDR}
Lior Yariv, Yoni Kasten, Dror Moran, Meirav Galun, Matan Atzmon, Basri Ronen,
  and Yaron Lipman.
\newblock Multiview neural surface reconstruction by disentangling geometry and
  appearance.
\newblock {\em NIPS}, 33, 2020.

\bibitem{yu2021artificial}
Ning Yu, Vladislav Skripniuk, Sahar Abdelnabi, and Mario Fritz.
\newblock Artificial fingerprinting for generative models: Rooting deepfake
  attribution in training data.
\newblock In {\em ICCV}, 2021.

\bibitem{kaizhang2020}
Kai Zhang, Gernot Riegler, Noah Snavely, and Vladlen Koltun.
\newblock Nerf++: Analyzing and improving neural radiance fields.
\newblock {\em arXiv:2010.07492}, 2020.

\bibitem{Zhang-2020-NAA}
Kai Zhang, Gernot Riegler, Noah Snavely, and Vladlen Koltun.
\newblock Nerf++: Analyzing and improving neural radiance fields.
\newblock {\em arXiv:2010.07492}, 2020.

\bibitem{zhang2018perceptual}
Richard Zhang, Phillip Isola, Alexei~A Efros, Eli Shechtman, and Oliver Wang.
\newblock The unreasonable effectiveness of deep features as a perceptual
  metric.
\newblock In {\em CVPR}, 2018.

\bibitem{CycleGAN2017}
Jun-Yan Zhu, Taesung Park, Phillip Isola, and Alexei~A Efros.
\newblock Unpaired image-to-image translation using cycle-consistent
  adversarial networks.
\newblock In {\em ICCV}, 2017.

\end{thebibliography}
}

\newpage

\begingroup
    \twocolumn[\centering\fontsize{16pt}{16pt}\selectfont
        \textbf{\MethodName: Fully Controllable Neural 3D Portraits}\\
 - \textbf{Supplementary} -\\ \vspace{1cm}]
\endgroup

\section{Ablations}

\begin{figure*}[h]
    \centering
    \includegraphics[width=\linewidth]{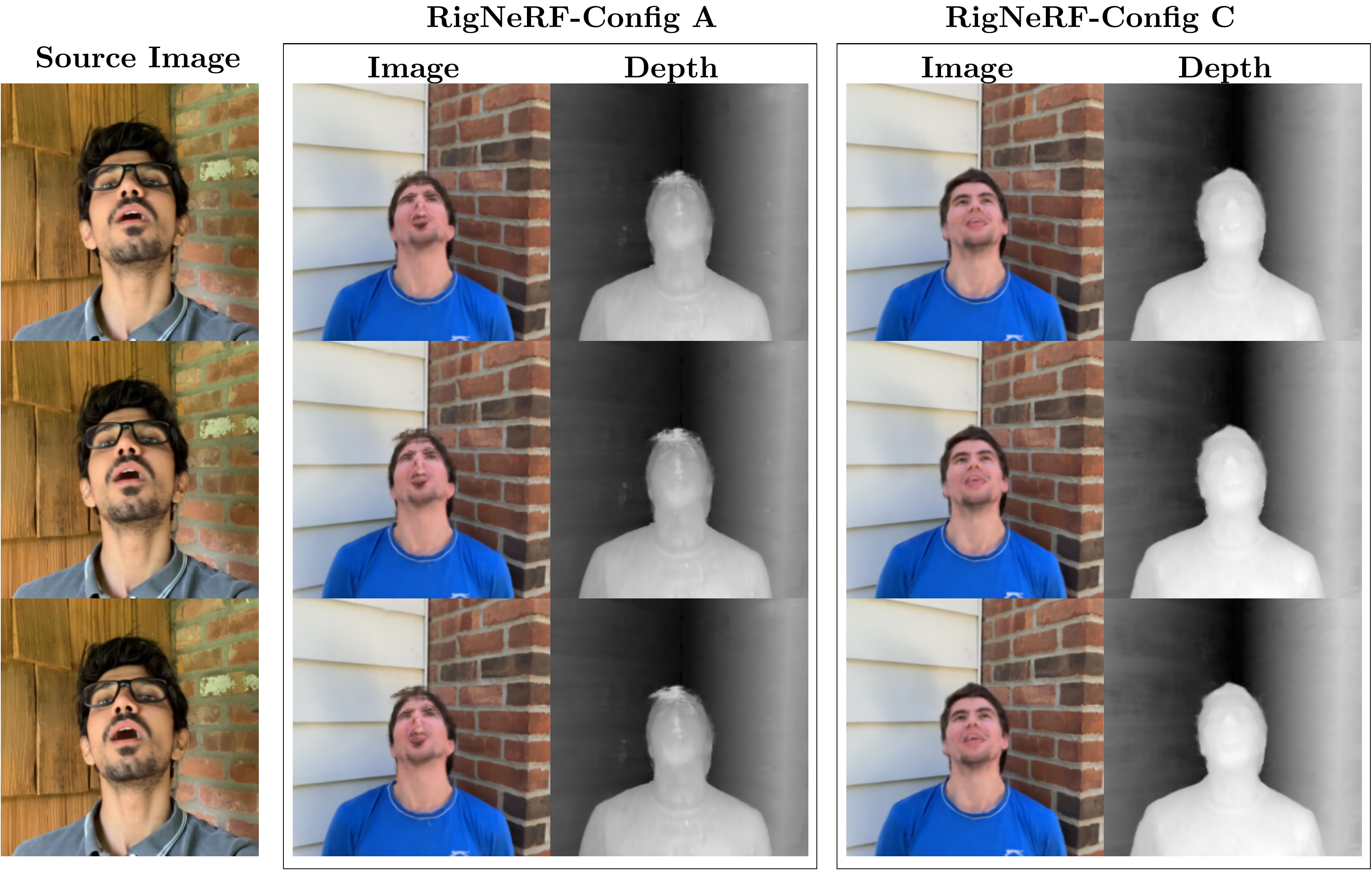}
    
    \caption{{\textbf{Qualitative comparison between \MethodName (Config A) and \MethodName (Config C)}
    }}
    \label{fig:aba-qual}
\end{figure*}

\begin{figure*}[h]
    \centering
    \includegraphics[width=\linewidth]{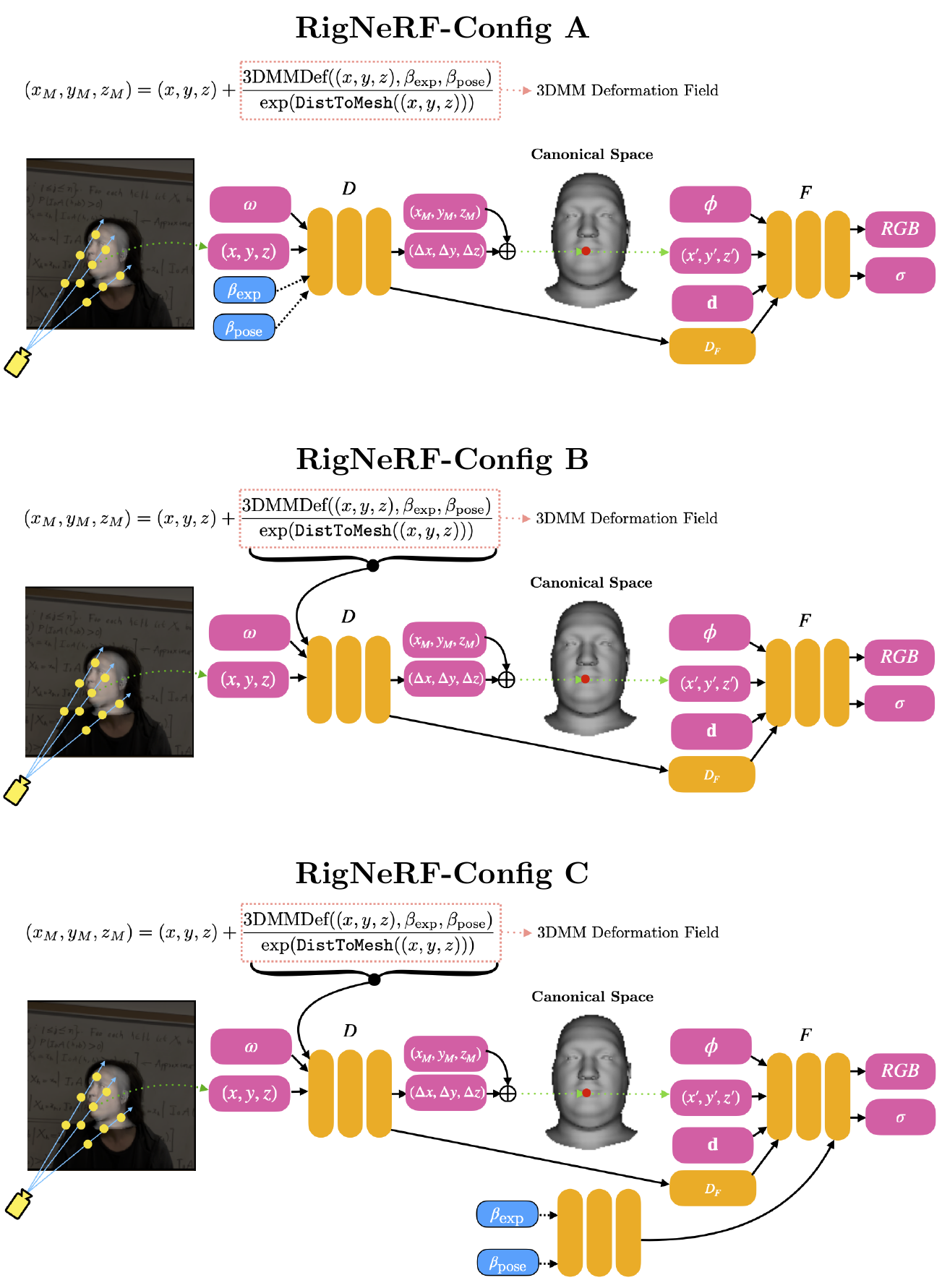}
    
    \caption{{\textbf{The various model configurations of \MethodName}
    }}
    \label{fig:model-configs}
\end{figure*}

In this section we ablate the various architectural choices of RigNeRF.  
\begin{itemize}
    \item{RigNeRF-Config A uses the most straightforward way to condition \(D\) on the 3DMM expression and pose parameters, \(\{\beta_{\text{exp}}, \beta_{\text{pose}}\}\), by providing them as input to \(D\). It does \textit{not} use a 3DMM conditioned appearance.}
    \item{RigNeRF-Config B tries to solve the overfitting of RigNeRF-Config A by using the 3DMM-deformation as input instead of the 3DMM expression and pose parameters. It does \textit{not} use a 3DMM conditioned appearance.}
    \item{RigNeRF-Config C uses the 3DMM-deformation as input and \textit{uses} a 3DMM-conditioned appearance, this is the final model used in the paper.}
\end{itemize}
\noindent A schema of all the configurations can be found in \fig{model-configs}. We show quantitative evaluations on the held-out test set in \tab{aba}. As can be seen, removing the 3DMM-parameters as input to \(D\) (Config A \(\rightarrow\) Config B) improves results, especially in terms of LPIPS and FaceMSE. Similarly, conditioning the appearance on 3DMM-parameters (Config B \(\rightarrow\) Config C), further improves these metrics. In \fig{aba-qual}, we show an example demonstrating poor generalization of Config A. Looking at both the final render generated by Config A and its depth, we see that it generalizes poorly to this novel head-pose. In contrast, the final model using Config C generalizes well to this novel headpose.

\begin{table}[h]
\centering
\begin{tabular}{@{}llll@{}}
\toprule
Method                     & PSNR  (\(\uparrow\))     & LPIPS  (\(\downarrow\))  & FaceMSE  (\(\downarrow\)) \\ \midrule
\MethodName-Config A & 28.5 & 0.183 & 1.5e-4 \\
\MethodName-Config B & 29.0 & 0.161 & 1.3e-4  \\ 
\MethodName-Config C & 29.45 & 0.119 & 9.8e-5 \\ \bottomrule

\end{tabular}
\caption{Ablation for model architecture.}
\label{tab:aba}
\end{table}

\section{FOMM Qualitative Results}
\begin{figure*}[h]
    \centering
    \includegraphics[width=\linewidth]{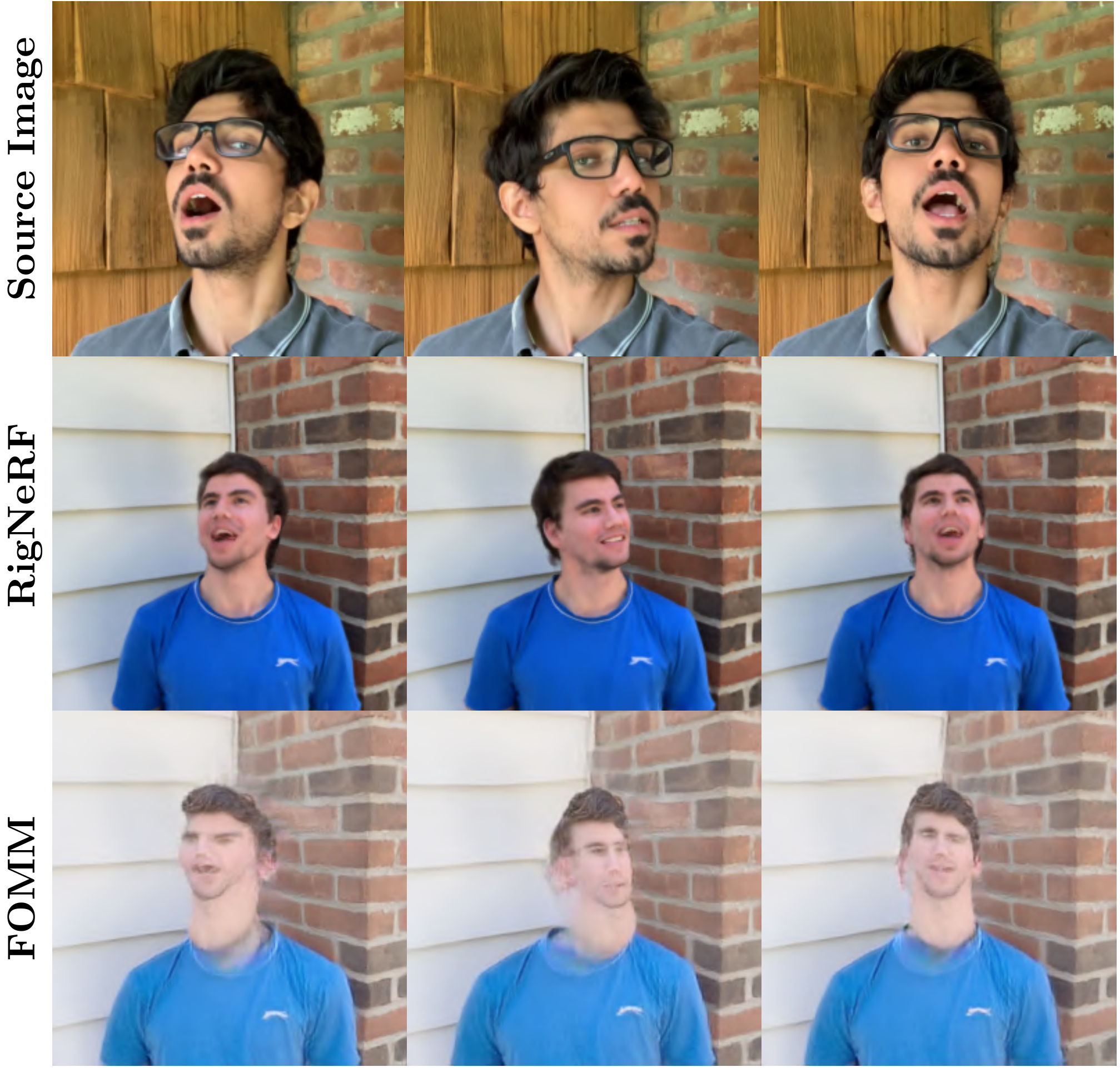}
    
    \caption{{\textbf{Qualitative comparison to FOMM \cite{siarohin2020first}} Here we reanimate \MethodName and FOMM \cite{siarohin2020first} using facial expression and head-pose derived from source images (top-row). The first row is the source image from which we transfer the facial expressions and head-pose, the second row shows the results of \MethodName while the third row shows the results of FOMM \cite{siarohin2020first}. We observe that FOMM generates significant artefatcs while \MethodName is able to reanimate the subject with high fidelity to the source facial expressions and head-pose.
    }}
    \label{fig:fomm_comp}
\end{figure*}

In \fig{fomm_comp}, we show a qualitative comparison between \MethodName  and FOMM \cite{siarohin2020first} as they're reanimated. We see that FOMM \cite{siarohin2020first} is unable to model the large deformations due to head-pose and facial expressions while \MethodName is able to do so with high-fidelity. 

\begin{figure*}[h]
    \centering
    \includegraphics[width=\linewidth]{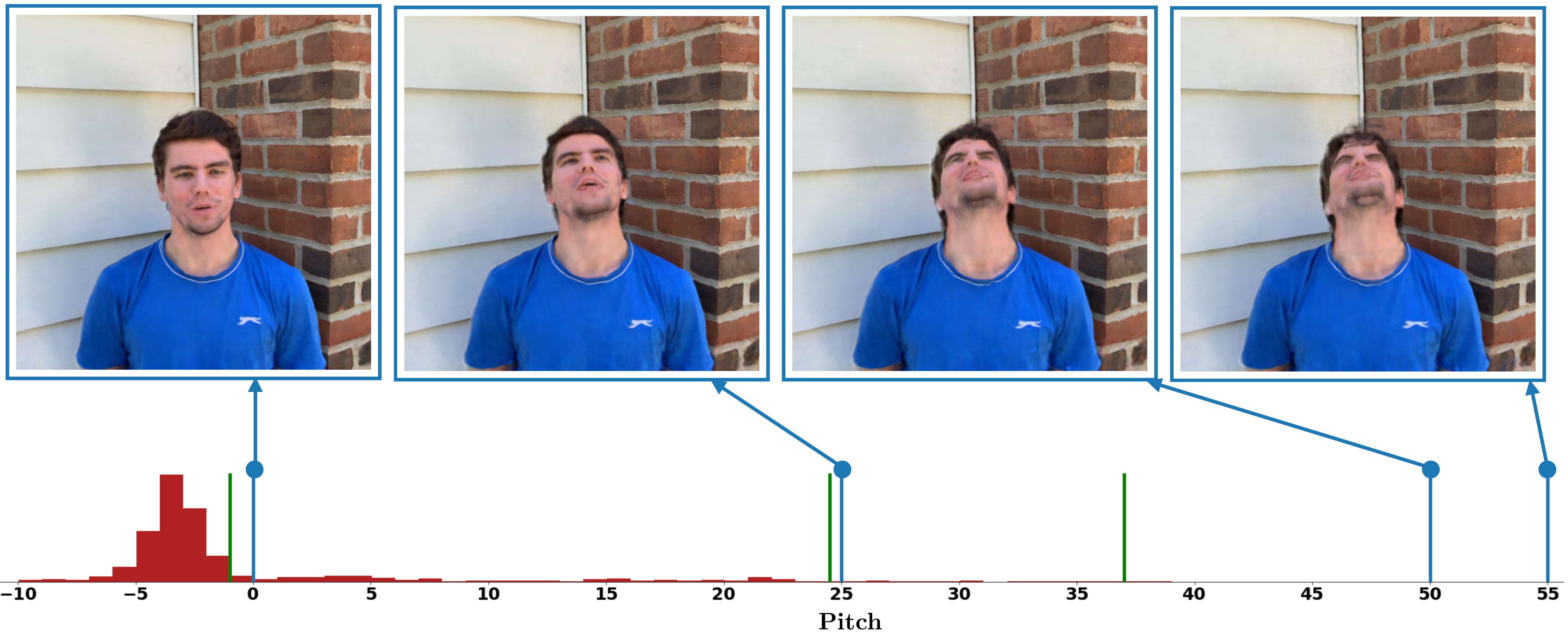}
    \vspace{-0.8cm}
    \caption{\footnotesize{{\textbf{Extrapolation} At the bottom we show a histogram of the pitch of the training data. In the top row we show reanimation results at pitch values sampled at the \textcolor{boldblue}{\textbf{bold blue lines.}} In \textcolor{boldgreen}{\textbf{green}} we show the closest training pitch to each image.   
    }}}
    \label{fig:pitch-extra}
\end{figure*}

\begin{table}[h]
\begin{center}
\small
\scalebox{0.7}{
\begin{tabular}{p{15mm}p{12mm}p{12mm}p{12mm}p{12mm}p{12mm}p{12mm}}
\toprule
  \multicolumn{7}{c}{\textbf{Head Yaw Generalization}} \\
  \midrule
  \(\Delta\)Angle   & -15.0 & -10.0 & -5.0  
   & 5.0 & 10.0 & 15.0\\
\midrule
 \textbf{LPIPS} (\(\downarrow\)) &
 0.174 & 0.156 & 0.131 &
 0.112 & 0.140 & 0.183 
 \\
\bottomrule
\end{tabular}}
\caption{Head Yaw (\(\Delta\)Angle to closest train yaw)}

\label{tab:Yaw}
\end{center}
\end{table}

\begin{table}[h]
\begin{center}
\small
\scalebox{0.8}{
\begin{tabular}{p{16mm}p{16mm}p{16mm}p{16mm}p{16mm}}
\toprule
   \multicolumn{5}{c}{\textbf{ Head Pitch Generalization}} \\
  \midrule
  \(\Delta\)Angle & 0.0 & 5.0 & 10.0 & 15.0\\
\midrule
 \textbf{LPIPS} (\(\downarrow\)) &
 0.100 & 0.111 & 0.125 &
 0.163 
 \\
\bottomrule
\end{tabular}}
\caption{Head Pitch (\(\Delta\)Angle to closest train pitch)}

\label{tab:Pitch}
\end{center}
\end{table}

\begin{table}[h]
\begin{center}
\small
\scalebox{0.8}{
\begin{tabular}{p{16mm}p{16mm}p{16mm}p{16mm}p{16mm}}
\toprule
  \multicolumn{5}{c}{\textbf{Jaw Pose Genralization}} \\
  \midrule
  \(\Delta\)Angle   & 0.0  & 10.0 & 15.0 & 20.0\\
\midrule
 \textbf{LPIPS} (\(\downarrow\)) &
 0.110 & 0.115 & 0.120 &
 0.125 
 \\
\bottomrule
\end{tabular}}
\caption{Jaw Pose (\(\Delta\)Angle to closest train jaw-pose)}

\label{tab:jawpose}
\end{center}
\end{table}

\begin{table}[h]
\begin{center}
\small
\scalebox{0.8}{
\begin{tabular}{p{16mm}p{16mm}p{16mm}p{16mm}p{16mm}}
\toprule
  \multicolumn{5}{c}{\textbf{Expression Generalization}} \\
  \midrule
  L2     & 0.001 &  0.010 & 0.100 & 0.340\\
\midrule
 \textbf{LPIPS} (\(\downarrow\)) &
 0.09 & 0.113 & 0.121 &
 0.152 
 \\
\bottomrule
\end{tabular}}
\caption{Expression Generalization (L2 to closest expression)}

\label{tab:exp}
\end{center}
\end{table}

\section{Generalization to Novel pose and expression}
In \fig{pitch-extra}, we show qualitative results of extrapolation of pitch outside the training data. In \tab{Yaw} and \tab{Pitch}, we show quantitative evaluations of extrapolation in head-pose yaw and pitch angles by measuring the perceptual distance to the closest frame. In \tab{jawpose} and \tab{exp}, we show results on jaw-pose and expression extrapolation. We will add this analysis to the paper. Additionally, we would like to note that Figs. 1, 4, 5 6 and 7 of the paper all show the results of reanimation on both novel head-poses and facial expressions.

\section{Experimental Details}

All our models were trained on 4 V100 GPUs using 128 samples per ray and a batch-size of 1550 rays. For all methods, we use 10 frequencies to encode the position of a point and 4 frequencies to encode the direction of the ray as  input to \(F\). For \MethodName and HyperNeRF,  we use 10 frequencies to encode the position of a point as input to \(D\). We also use a coarse-to-fine regularization on the positional encoding of the position input to \(D\) for 40k epochs. For the ambient dimensions used in HyperNeRF \cite{park2021hypernerf}, we follow the same annealing schedule as specified in the paper. The architecture of \(F\) is identical to the original NeRF MLP \cite{nerf} with 8 layers of 256 hidden units each. We use the same \(F\) for all our methods. \(D\) consists of 8 layers with 128 hidden units in each. We use the same architecture for \(D\) for both \MethodName and HyperNeRF. We train \MethodName and NerFACE with an initial learning rate of \(5e^{-4}\) which is decayed to \(5e^{-5}\) by the end of training. For HyperNeRF \cite{park2021hypernerf}, we use an initial learning rate of \(1e^{-3}\) which  is decayed to \(1e^{-4}\) by the end of training. Details about the training epochs and the dimensionality of the appearance and deformation codes used is given in \tab{model-details}

\begin{table*}[h]
\centering
\begin{tabular}{@{}lllll@{}}
\toprule
Subject                     &Method                     & Epochs Trained       & App Code dim  & Def Code dim  \\ \midrule
Subject 1 &\MethodName &150000 & 8 & 8 \\
&HyperNeRF                   & 150000  & 8 & 64 \\ 
&NerFACE                   & 150000  & 32 & N/A \\ \bottomrule
Subject 2 &\MethodName &150000 & 8 & 8 \\
&HyperNeRF                   & 150000  & 8 & 64 \\ 
&NerFACE                   & 150000  & 32 & N/A \\ \bottomrule
Subject 3 &\MethodName &100000 & 8 & 8 \\
&HyperNeRF                   & 100000  & 8 & 64 \\ 
&NerFACE                   & 100000  & 32 & N/A \\ \bottomrule
Subject 4 &\MethodName &100000 & 8 & 8 \\
&HyperNeRF                   & 100000  & 8 & 64 \\ 
&NerFACE                   & 100000  & 32 & N/A \\ \bottomrule
\end{tabular}
\caption{Training configuration for all the experiments.}
\label{tab:model-details}
\end{table*}

\section{Societal Impact}
Being a method that is capable of reanimating faces, \MethodName is prone to misuse by bad actors to generate deep-fakes. However, work such as \cite{wang2019cnngenerated} has shown that it is possible to train discriminative classifiers to detect images and videos generated by synthetic methods like ours. Other possible solution include watermarking training images \cite{yu2021artificial} so that they can be more easily detected.

\end{document}